%% file: main.tex
\pdfoutput=1
\documentclass[11pt]{article}

\usepackage[final]{acl}

\usepackage{times}
\usepackage{latexsym}

\usepackage[T1]{fontenc}
\usepackage[utf8]{inputenc}

\usepackage{microtype}

\usepackage{inconsolata}

\usepackage{graphicx}
\usepackage{xcolor}
\definecolor{mygray}{HTML}{404040}

\input{header}

\input{math_commands}

\title{\quad\quad Diagnosing Memorization in Chain-of-Thought Reasoning\newline One Token at a Time}

\author{Huihan Li$^{1*}$ \hspace{1mm} You Chen$^{2*}$ \hspace{1mm} Siyuan Wang$^{1}$ \hspace{1mm} \textbf{Yixin He}$^{1}$ \hspace{1mm} Ninareh Mehrabi$^{3\dag}$ \hspace{1mm} \\ \textbf{Rahul Gupta}$^{4}$ 
\hspace{1mm} \textbf{Xiang Ren}$^{1}$\\
$^{1}$University of Southern California \hspace{1mm}
$^{2}$University of California, San Diego\\
$^{3}$Meta \hspace{1mm} $^{4}$Amazon AGI\\
\small{\texttt{\{huihanl,xiangren\}@usc.edu, yoc033@ucsd.edu}}}

\begin{document}
\maketitle
\renewcommand{\thefootnote}{\fnsymbol{footnote}}
\footnotetext[1]{Equal Contribution}
\footnotetext[2]{Work done in Amazon AGI}
\renewcommand{\thefootnote}{\arabic{footnote}}
\begin{abstract}
Large Language Models (LLMs) perform well on reasoning benchmarks but often fail when inputs alter slightly, raising concerns about the extent to which their success relies on memorization. This issue is especially acute in Chain-of-Thought (CoT) reasoning, where spurious memorized patterns can trigger intermediate errors that cascade into incorrect final answers. We introduce \framework, a novel framework
for \textbf{S}ource-aware \textbf{T}oken-level \textbf{I}dentification of \textbf{M}emorization, which attributes each token in a reasoning chain to one of multiple memorization sources -- local, mid-range, or long-range -- based on their statistical co-occurrence with the token in the pretraining corpus. Our token-level analysis across tasks and distributional settings reveals that models rely more on memorization in complex or long-tail cases, and that local memorization is often the dominant driver of errors, leading to up to 67\% of wrong tokens. We also show that memorization scores from \framework\ can be effective in predicting the wrong tokens in the wrong reasoning step. \framework\ offers a powerful tool for diagnosing and improving model reasoning and can generalize to other structured step-wise generation tasks.\footnote{\url{https://github.com/INK-USC/STIM}}
\end{abstract}

\input{sections/intro}
\input{sections/related}

\input{sections/methodology}
\input{sections/analyses}
\input{sections/disc_conclusion}

\newpage

\input{sections/limitations}

\section*{Acknowledgments}
This research is supported in part by the Office of the Director of National Intelligence (ODNI), Intelligence Advanced Research Projects Activity (IARPA), via the HIATUS Program contract \#2022-22072200006, the Defense Advanced Research Projects Agency with award HR00112220046, and NSF IIS 2048211. The views and conclusions contained herein are those of the authors and should not be interpreted as necessarily representing the official policies, either expressed or implied, of ODNI, IARPA, or the U.S. Government. Huihan Li was supported by Amazon ML Fellowship from the USC + Amazon Center on Secure and Trusted Machine Learning.

\bibliography{custom}

\newpage

\appendix

\input{sections/appendix}

\end{document}

%% file: header.tex
\usepackage{booktabs}
\usepackage{url}
\usepackage{times}
\usepackage{latexsym}

\usepackage{amsmath,amssymb}
\usepackage{graphicx}
\usepackage{adjustbox}
\usepackage{tabularx}
\usepackage{ragged2e}
\usepackage{multirow}
\usepackage{pifont}
\usepackage{soul}

\usepackage[toc,page]{appendix}
\usepackage{xcolor}
\usepackage{color}
\usepackage{floatrow}
\usepackage{makecell}
\usepackage{xspace}
\usepackage{colortbl}
\usepackage{enumitem}
\usepackage[most]{tcolorbox}

\definecolor{airforceblue}{rgb}{0.36, 0.54, 0.66}
\definecolor{ao(english)}{rgb}{0.0, 0.5, 0.0}
\definecolor{auburn}{rgb}{0.43, 0.21, 0.1}
\definecolor{bittersweet}{rgb}{1.0, 0.44, 0.37}

\usepackage[tableposition=top]{caption}
\usepackage{floatrow}
\usepackage{subcaption}
\usepackage{wrapfig}
\usepackage{makecell}

\newfloatcommand{capbtabbox}{table}[][\FBwidth]

\usepackage{arydshln}

\usepackage{minibox}
\usepackage{tabularx,ragged2e}
\usepackage{tabularx} % in the preamble
\usepackage{algorithm}
\usepackage[noend]{algpseudocode}
\usepackage{bbm}

\newcommand{\framework}{\textbf{STIM}}

\newcommand{\olmo}{\texttt{OLMo-2-1124-13B-Instruct}}

\newcommand{\cotr}{Chain-of-thought}

%% file: math_commands.tex
\usepackage{amsmath,amsfonts,bm}

\def\Figref#1{Figure~\ref{#1}}

\def\Secref#1{Section~\ref{#1}}
\def\eqref#1{equation~\ref{#1}}
\def\Tabref#1{Table~\ref{#1}}

\def\1{\bm{1}}

\DeclareMathAlphabet{\mathsfit}{\encodingdefault}{\sfdefault}{m}{sl}
\SetMathAlphabet{\mathsfit}{bold}{\encodingdefault}{\sfdefault}{bx}{n}

%% file: sections/intro.tex
\section{Introduction}
Large Language Models (LLMs) perform well on reasoning tasks but often fail under slight input changes, raising concerns about overreliance on memorization~\citep{hong2025reasoning,lou2024quantifying,jin2024disentangling,salido2025none}. Long \cotr\ (CoT)~\citep{wei2022chain} chains are especially vulnerable, as spurious memorization can introduce early errors that derail final answers. As inference-time scaling encourages longer CoTs, detecting token-level memorization is critical for assessing reasoning reliability, particularly under distributional shifts from frequent to rare inputs~\citep{xie2024memorization,prabhakar2024deciphering}.
\looseness=-1

\input{figures/motivation}

We argue that memorization in long \cotr\ generations must be identified at the \textbf{token level} rather than the sequence level. A single faulty step can cause cascading errors, often stemming from a few erroneous tokens (\Figref{fig:motivation}). Identifying these tokens and whether they result from memorization is essential. Moreover, we argue that accurately measuring token-level memorization requires accounting for \textbf{multiple sources of influence}, including both the input prompt and prior output tokens, which jointly shape each token’s generation (\Tabref{tab:dominant_sources}).

\input{figures/figure1}

\input{tables/different_sources}

Prior approaches are insufficient for analyzing token-level memorization and how memorization patterns from different sources shift under distributional change: existing metrics do not target memorization at the level of individual tokens, instead reporting a single score for the entire sequence or final answer. Moreover, they either focus solely on memorization in the output sequence~\citep{mccoy2023much,merrill2024evaluating,lu2024ai} or assess the influence of memorization from the input~\citep{carlini2022quantifying,biderman2023emergent,liattributing,wang2024generalization}, without accounting for multiple sources of influence on token-level memorization.
\looseness=-1

To address these gaps, we propose \textbf{STIM} (\textbf{S}ource-aware \textbf{T}oken-level \textbf{I}dentification of \textbf{M}emorization), a framework that captures token-level memorization by tracing influences from both the input and prior outputs on erroneous reasoning steps. For each token, \framework\ computes the strength of three memorization sources: (1) \textit{local}, from frequent continuations of immediately preceding tokens; (2) \textit{long-range}, from frequent co-occurrence with prompt tokens; and (3) \textit{mid-range}, when the model generates the target token when conditioned only on a prefix of the generation, we identify tokens that frequently co-occur with the target token in pretraining. \framework\ offers a fine-grained view of multi-source memorization and its strength at each token.

We begin our analysis by using \framework\ to uncover broad memorization trends across tasks, input distributions, and correctness. As reasoning complexity increases, models exhibit greater reliance on memorization. Distribution shifts toward rare or atypical inputs also lead to stronger memorization signals. Interestingly, while memorization often supports correct answers in base settings, it more frequently contributes to errors in long-tail scenarios, suggesting defective recall when faced with unfamiliar contexts.

To demonstrate the utility of our framework, we apply it to the task of identifying erroneous tokens in erroneous reasoning steps. By tracing the dominant source of memorization for each incorrect token, we find that local memorization (continuations driven by immediately preceding tokens) is the most common cause of error (up to 67\%). However, under distribution shift, complex tasks show a marked decline in local memorization-driven mistakes, implying reduced reliance on familiar patterns. Finally, we assess the effectiveness of \framework\ in pinpointing erroneous tokens via Precision@k and Recall@k, showing that high memorization scores are strong indicators of reasoning failures.
\looseness=-1

Our novel \framework\ framework is the first to enable fine-grained, token-level memorization identification in long reasoning chains, considering multiple influence sources from both input and generated context. Through systematic evaluation across diverse \cotr\ tasks, we demonstrate its crucial role in pinpointing memorized content that leads to errors and reveal how different memorization sources dominate across tasks and under distributional shift. While focused on \cotr, \framework\ is extendable to other long-form formats such as dialogue or summarization. This framework offers a powerful tool for diagnosing LLM reasoning failures and advancing research into genuine reasoning capabilities.

%% file: figures/motivation.tex
\begin{figure}[t]
    \centering
    \includegraphics[width=\textwidth]{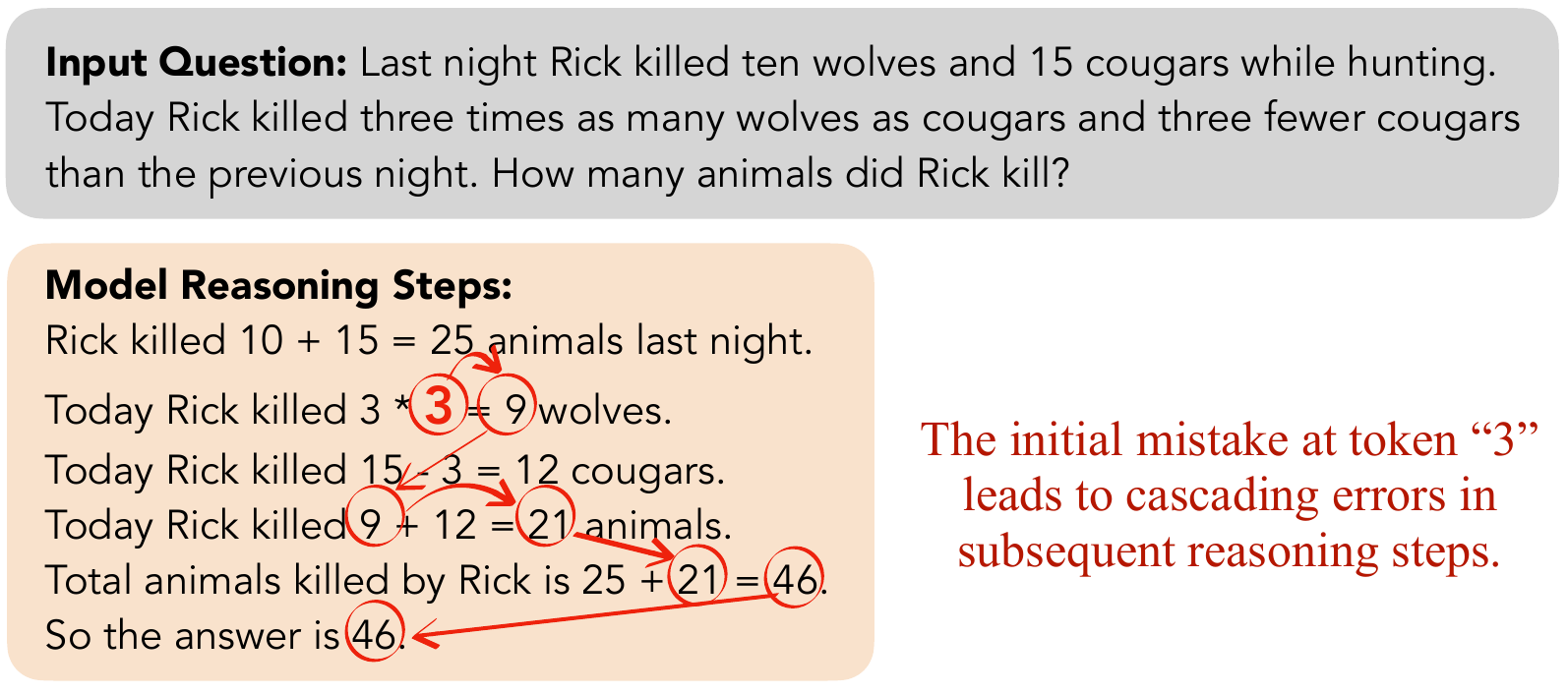}
    \caption{Cascading errors in \cotr\ (CoT) reasoning can stem from a single mis-predicted token, often influenced by incorrect memorization of pretraining data. This motivates our study on the impact of token-level memorization.
    \vspace{-2em}
    }
    \label{fig:motivation}
\end{figure}

%% file: figures/figure1.tex
\begin{figure}[t]
    \centering
    \includegraphics[width=\linewidth]{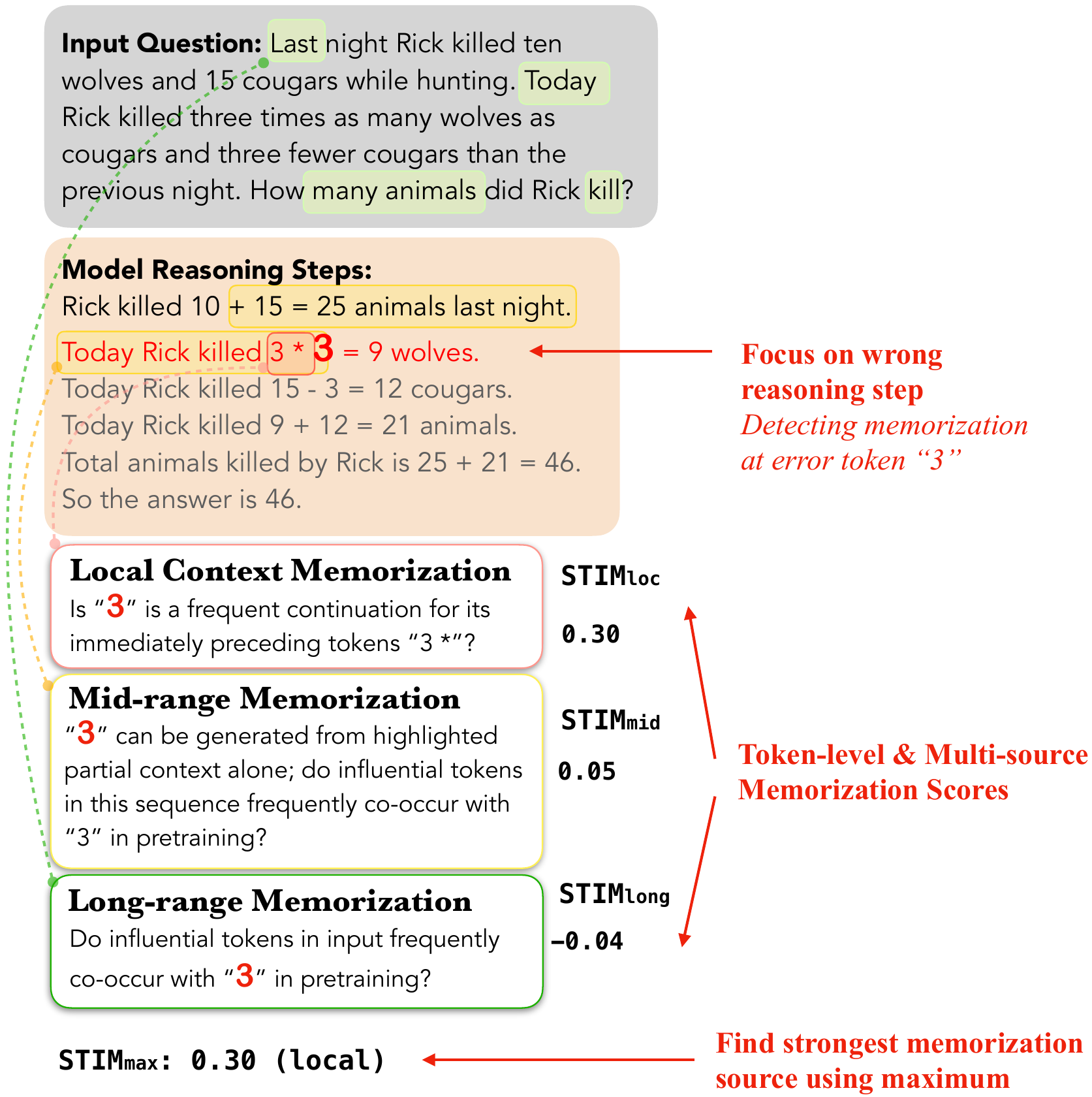}
    \caption{\framework\ quantifies local, mid-range, and long-range memorization at each token in a faulty reasoning step and identifies the dominant source. \framework\ is helpful for detecting error tokens.
    \vspace{-1.5em}
    }
    \label{fig:figure1}
\end{figure}

%% file: tables/different_sources.tex
% \begin{table*}[h]
% \centering
% \small
% \begin{tabularx}{\linewidth}{X}
% \toprule
% \textbf{Examples} \\
% \midrule
% \textbf{Question}: Here is a list: [pitanga, pitanga, yumberry, yumberry, pitanga, yumberry, pitanga, yumberry, pitanga, pitanga]. How many times does 'pitanga' appear on it? \newline
% \textbf{Model output}: \newline
% 1. The first element is 'pitanga'. \newline
% ... \newline
% 10. The tenth element is 'pitanga'. \newline
% Now, counting the occurrences of 'pitanga':- We have 'pitanga' appearing 5 times at positions 1, 2, 5, 7, and 10. \newline
% So the answer is 5. \newline
% \textbf{Analysis}: Token “5” is wrong, and the memorization score is (local=0.81, mid=0.24, long=0.03). The longest non-zero sequence is ``appearing 5'', which has a higher frequency.
% \\
% \midrule
% \textbf{Question}: What is 13817 + (13817 / 247 + 11.67 * 13817 / 247) * 247 * 2 equal to? \newline
% \textbf{Model Output}: \newline
% To solve this, first calculate 13817 / 247 which equals 55. \newline
% ... \newline
% So the answer is 10271737. \newline
% \textbf{Analysis}: token 55 is wrong and it has a high mid memorization score (local=0.168, mid=0.528, long=0.159). \newline
% Shortest prefix is “13817 / 247 which equals” and selected words like “13817, 247” have high co-occurrence frequency with 55.
% \\
% \bottomrule
% \end{tabularx}
% \caption{Examples of different dominant memorization sources.}
% \label{tab:dominant_sources}
% \end{table*}

\definecolor{highlight}{HTML}{D9EDF7} % Light blue for dominant source

\begin{table*}[h]
\centering
\small
\begin{tabularx}{\linewidth}{X}
\toprule
\textbf{Examples of Simultaneous Memorization Sources of Different Strength} \\
\midrule
\textbf{Question}: Here is a list: [pitanga, pitanga, yumberry, yumberry, pitanga, yumberry, pitanga, yumberry, pitanga, pitanga]. How many times does 'pitanga' appear on it? \newline

\textbf{Model Output}: \newline
1. The first element is 'pitanga'. \newline
... \newline
10. The tenth element is 'pitanga'. \newline
Now, counting the occurrences of 'pitanga': We have 'pitanga' \colorbox{orange}{appearing} \colorbox{highlight}{5} times at positions 1, 2, 5, 7, and 10. \newline
So the answer is 5. \\
\midrule
\textbf{Analysis}: Token \colorbox{highlight}{5} is incorrect and has the highest influence from local context, where the n-gram ``appearing 5'' has a high pretraining frequency. \framework\ scores are: \colorbox{orange}{local = 0.81}, mid = 0.24, long = 0.03, suggesting that the model was influenced by locally frequent patterns rather than actual counting.
\\
\bottomrule
\toprule
\textbf{Question}: What is (\colorbox{pink}{100101100} - 100101100 * 0.1) * 111110100 + 100101100 * 1111 / 1100100 * 111110100 \colorbox{pink}{*} 0.1 + (100101100 - 100101100 * 0.1 - \colorbox{pink}{100101100} \colorbox{pink}{*} 1111 / 1100100) * \colorbox{pink}{111110100} equal to? (base-2) \newline

\textbf{Model Output}: \newline
To solve this, we first handle the multiplication and division in base-2, converting to base-10 as necessary: \newline
1. $(100101100 \times 0.1_{\text{base-2}} = 10010110_{\text{base-2}})$ (which is \colorbox{highlight}{180} in base-10) \newline
... \newline
So the answer is 26807.536.
\\
\midrule
\textbf{Analysis}: Token \colorbox{highlight}{180} is incorrect and has the highest influence from input tokens, having higher pretraining co-occurrence frequency with “111110100”, “*” and “100101100” in the input. \framework\ scores are local=-0.19, mid=0.09, \colorbox{pink}{long=0.156}, indicating long-range memorization being the primary influence.
\\
\bottomrule
\end{tabularx}
\caption{Each \colorbox{highlight}{error token} is affected simultaneously by different memorization sources, but some sources may be stronger than others. Shown are examples with local context (\colorbox{orange}{local}) and input prompt (\colorbox{pink}{long}) having maximum influence, with tokens associated with the source highlighted. How we obtain \framework\ scores for each source is explained in~\Secref{subsec:stim}.}

\label{tab:dominant_sources}
\end{table*}

%% file: sections/related.tex
\section{Related Works}

\paragraph{Pretraining Data Memorization Metrics.}
A growing body of work investigates how and when large language models (LLMs) memorize their pretraining data. Early approaches focused on measuring extractability -- the ease with which specific training data can be reproduced from the model—highlighting risks to privacy and model overfitting~\citep{carlini2022quantifying,biderman2023emergent}. Other efforts have analyzed memorization via novelty metrics, which assess the similarity of model outputs to seen data~\citep{mccoy2023much,merrill2024evaluating,lu2024ai}. Another class of work quantifies memorization through n-gram overlap or token-level attribution, often drawing connections between model predictions and local or distant pretraining contexts~\citep{liattributing,wang2024generalization}.

\paragraph{Distinguishing Memorization from Reasoning.}
Beyond surface-level memorization, recent work has aimed to disentangle genuine reasoning from pattern recall. \citet{xie2024memorization} fine-tune models on controlled datasets to probe the boundary between memorization and generalization in reasoning tasks. Complementary to this, \citet{hong2025reasoning} approach the problem from a mechanistic interpretability perspective, identifying internal circuits associated with memorized versus reasoned responses. On the modeling side, \citet{lou2024quantifying} propose methods for quantifying chaotic and foundational memorization by analyzing in-context learning dynamics at the logit level, while \citet{jin2024disentangling} focus on how training conditions shape these behaviors. In multiple-choice settings, \citet{salido2025none} introduce a technique to isolate reasoning by deliberately eliminating answer options that could be matched through memorized heuristics. Finally, \citet{prabhakar2024deciphering} study the interplay between probability calibration, memorization, and noise in chain-of-thought prompting, revealing that correct answers can arise from superficial token-level cues rather than coherent reasoning.
\looseness=-1

%% file: sections/methodology.tex
% \section{Measuring Token-wise Memorization in \cotr\ Reasoning}

% \section{Analyzing Distribution Shift Impact on CoT Reasoning}
\section{Preliminary Analysis: CoT Reasoning-Memorization Correlation}

Recent work shows that language models struggle to generalize to rare task formats or uncommon input entities despite strong performance on frequent patterns, suggesting that memorization significantly influences model behavior~\citep{wu2024reasoning,dziri2023faith,li2024search}. To investigate how memorization impacts CoT Reasoning, we construct a controlled set of reasoning tasks and introduce targeted long-tail transformations that either decrease the frequency of input entities or alter task formats to less common variants.
This section introduces our experimental setup and motivates our framework by demonstrating how reasoning performance changes under distributional shift.

\subsection{Setup}

\paragraph{Tasks.} We evaluate memorization on four reasoning tasks of varying complexity: \textit{Applied Math, Formula Calculation, Counting, and Capitalization.} All tasks require step-by-step reasoning under a \cotr\ prompting setup as well as a direct answer setup(full prompt in Appendix~\ref{app:prompt}). We collect/construct each task as follows (more details in Appendix~\ref{app:data_construction}:

\begin{itemize}[itemsep=-4pt, topsep=0pt]
    \item \textit{Applied Math}: GSM8K~\citep{cobbe2021training}.
    \item \textit{Formula Calculation}: Final equations extracted from GSM8K answer derivations.
    \item \textit{Counting}: Lists of fruit entities with controlled frequency ranges ($10^3$ to $10^7$) and varying lengths ($10$–$50$).
    \item \textit{Capitalization}: Book titles sampled from the Book Cover Dataset~\citep{iwana2016judging}.
\end{itemize}

These tasks involve different reasoning complexity levels.  \textit{Applied Math} and \textit{Formula Calculation} involve multi-step deduction, where each intermediate step depends on correct synthesis of all prior steps. In contrast, \textit{Counting} and \textit{Capitalization} are more locally structured, where each step is relatively independent, and the correct answer often depends on a single, final decision. This distinction allows us to examine how memorization interacts with different reasoning dynamics, especially when tracing errors in multi-step generation.

\paragraph{Long-tail Transformations.}
To evaluate memorization under distributional shift, we apply long-tail transformations to each task (\Tabref{tab:head_lt_eg}). These modifications reduce the frequency of input entities or reformulate tasks in less common formats:

\begin{itemize}[itemsep=-4pt, topsep=0pt]
    \item \textit{Applied Math \& Formula Calculation}: Frequency reduction via digit expansion or converting integers to floats; task variation by expressing problems in base-2~\citep{li2024gsm}.
    \item \textit{Counting}: Lower-frequency countable entities and increased list lengths.
    \item \textit{Capitalization}: Atypical formulation requiring capitalization of the last word’s first letter.
\end{itemize}
Each transformation is applied independently, and the final long-tail set per task pools all variants. The total number of base and long-tail examples we collect is in~\Tabref{tab:num_examples}.

\paragraph{Model and Pretraining Data.}
Our experiments and analyses utilize OLMo 2\citep{olmo20242} (\olmo) under default HuggingFace settings and greedy decoding. Its pretraining corpus, Dolma 1.7~\citep{soldaini2024dolma}, is indexed via Infinigram~\citep{liuinfini}, allowing token frequency lookup. To our knowledge, only OLMo and Pythia offer fully open, indexed pretraining data; however, Pythia underperforms on GSM8K ($<$5\%), limiting its use in CoT studies. To show generalizability of our methodology, we replicate all main analyses on \texttt{olmo2-1124-7B-Instruct} in Appendix~\ref{app:reproducing}.

\subsection{Performance under Distribution Shift}

OLMo 2's performance across both direct answer and CoT formats for all tasks is listed in~\Tabref{tab:reasoning_accuracy_combined}.
\input{tables/accuracy}

\paragraph{Base Distribution Outperforms Long-tail}  
For \textit{Applied Math}, \textit{Formula Calculation}, and \textit{Counting}, performance in the base distribution setting consistently outperforms that in the long-tail distribution. This trend reinforces the notion that current models rely heavily on memorization, benefiting from high-frequency entities and familiar patterns in the input. The observed performance degradation under long-tail input distributions reflects the model's difficulty generalizing to rare entities or less common problem variations.

\paragraph{CoT Amplifies the Base vs. Long-tail Gap}
The performance gap between base and long-tail distributions is significantly larger under the CoT format compared to direct answers. This suggests that the extended generation sequences in CoT reasoning exacerbate the model’s reliance on memorized patterns, increasing error probability when familiar cues are absent. The phenomenon necessitates deeper analysis of how CoT prompts may amplify memorization-related errors.

\paragraph{Capitalization Highlights Token-level Fragility}  
An exception arises in the \textit{Capitalization} task, where CoT notably degrades performance in the base setting. While the model can directly retrieve correct answers for well-known book titles in the direct answer format, CoT reasoning introduces intermediate steps that expose the model to more opportunities for spurious generations. This example illustrates that while certain forms of memorization (e.g., direct recall) can support accurate performance, others (e.g., misplaced pattern matching during reasoning) can introduce critical errors. The discrepancy underscores the need for token-level analysis to understand when memorization helps versus when it harms.

In summary, the results demonstrate that memorization plays a central role in model behavior under distribution shift, particularly under CoT, where longer reasoning chains amplify its effects. This motivates our fine-grained analysis of token-level memorization dynamics in following sections.

\section{Measuring Token-Level Memorization}
% \section{Source-aware}
\label{subsec:stim}

In multi-step reasoning, token predictions are influenced by the local context, the input prompt, and previously generated output. Each contributes to memorization differently: local context drives frequent continuations, while prompts and past outputs reflect longer-range associations from pretraining. Disentangling these sources enables more precise diagnosis of how memorization affects reasoning, especially under distributional shifts. We introduce \textbf{S}ource-aware \textbf{T}oken-level \textbf{I}dentification of \textbf{M}emorization (\framework), a method for identifying token-level memorization from local, mid-range, and long-range sources.

\subsection{Memorization from Distinct Contextual Sources}

We now describe how to quantify memorization for a target token $x$, conditioned on the full context $p = [\text{input}; \text{output}_{<x}]$, where $\text{output}_{<x}$ denotes all preceding tokens in the generated answer before $x$.

\paragraph{Local Context Memorization (\textit{Local})}
This score quantifies how much a token’s generation is driven by frequent continuations in its immediate local context.
For a target token $x$, we identify $w$, the longest contiguous prefix such that the n-gram $[w; x]$ appears at least once in the pretraining corpus. At decoding time, we extract the top-20 candidate tokens ${x_i}_{i=1}^{20}$ with probabilities $P(x_i \mid p)$ and retrieve their corresponding n-gram frequencies $f([w; x_i])$. The local memorization score is computed as the Spearman correlation:

\vspace{-2ex}
\begin{equation}
\small
\text{STIM}_{loc}(x) = \rho\left( 
\{ P(x_i \mid p) \}_{i=1}^{20},\ 
\{ f([w; x_i]) \}_{i=1}^{20} 
\right)
\end{equation}

\paragraph{Input-driven Memorization (\textit{Long-Range})}
This score measures the influence of salient input tokens that co-occurred with the target token in pretraining.
We identify the top-5 most influential input tokens to the target token using token saliency~\citep{tuan2021local}\footnote{Ablation see Appendix~\ref{app:ablation_prm_lerg}.}, denoted $S_l = {s_1, \dots, s_5}$. At decoding time, we obtain the top-20 candidates ${x_i}$ with probabilities $P(x_i \mid p)$ and compute their co-occurrence frequencies with $S_l$, denoted $f(S_l, x_i)$. The long-range memorization score is defined as:

\vspace{-2ex}
\begin{equation}
\small
\text{STIM}_{long}(x) = \rho\left(
{ P(x_i \mid p) }_{i=1}^{20},\\
{ f(S_l, x_i) }_{i=1}^{20}
\right)
\end{equation}

\paragraph{Partial Output Memorization (\textit{Mid-Range})}
This score captures how much a token’s generation is influenced by spurious associations within the partially generated answer.
For a target token $x$, we find the shortest contiguous prefix of the model’s generated answer (excluding input tokens) that leads to the model generating $x$ when conditioned on that span.
We then identify the top-5 most salient tokens in this context $S_m = \{s_1, \dots, s_5\}$ using token saliency. We collect the top-20 candidate tokens $\{x_i\}_{i=1}^{20}$ with probabilities $P(x_i \mid p)$. For each candidate $x_i$, we compute the average co-occurrence frequency with the salient tokens $S$, denoted $f(S_m, x_i)$. The mid-range memorization score is defined as:

\vspace{-2ex}
\begin{equation}
\small
\text{STIM}_{mid}(x) = \rho\left(
{ P(x_i \mid p) }_{i=1}^{20},\\
{ f(S_m, x_i) }_{i=1}^{20}
\right)
\end{equation}

\paragraph{Dominant Source Attribution}
To capture the full extent of memorization effects at the token level, we also introduce $\text{STIM}_{max}$, taking the maximum of local, long and mid-range memorization scores. We identify the \textbf{dominant source} as the memorization source with the highest score, indicating which contextual factor most strongly influenced the token’s generation.

%% file: tables/accuracy.tex
% \begin{table*}[t]
% \centering
% \caption{Model \textbf{COT} accuracy (\%) on four reasoning tasks under head and long-tail input distributions.}
% \label{tab:reasoning_accuracy_cot}
% \begin{tabular}{lcc}
% \toprule
% \textbf{Task} & \textbf{Head} & \textbf{Long-tail} \\
% \midrule
% Applied Math & 82.0 & 25.6 \\
% Formula Calculation & 89.8 & 39.3 \\
% Counting & 43.1 & 19.2 \\
% Capitalization & 26.7 & 53.1 \\
% \bottomrule
% \end{tabular}
% \end{table*}

% \begin{table*}[t]
% \centering
% \caption{Model \textbf{direct} accuracy (\%) on three reasoning tasks (Applied math performance is too low) under head and long-tail input distributions.}
% \label{tab:reasoning_accuracy_std}
% \begin{tabular}{lcc}
% \toprule
% \textbf{Task} & \textbf{Head} & \textbf{Long-tail} \\
% \midrule
% Formula Calculation & 33.3 & 11.3 \\
% Counting & 40.5 & 21.4 \\
% Capitalization & 87.5 & 47.8 \\
% \bottomrule
% \end{tabular}
% \end{table*}

\begin{table}[!htb]
\centering
\caption{Model accuracy (\%) on reasoning tasks under base and long-tail input distributions. Direct Answer for Applied Math is omitted due to poor accuracy.}
\label{tab:reasoning_accuracy_combined}
\small
\begin{subtable}[t]{\textwidth}
\centering
\caption{\textbf{CoT Reasoning}}
\begin{tabular}{lcc}
\toprule
\textbf{Task} & \textbf{Base} & \textbf{Long-tail} \\
\midrule
Applied Math & 82.0 & 25.6 \\
Formula Calculation & 89.8 & 39.3 \\
Counting & 43.1 & 19.2 \\
Capitalization & 26.7 & 53.1 \\
\bottomrule
\end{tabular}
\label{tab:cot_prompting}
\end{subtable}

\begin{subtable}[t]{\textwidth}
\centering
\caption{\textbf{Direct Answer}}
\begin{tabular}{lcc}
\toprule
\textbf{Task} & \textbf{Base} & \textbf{Long-tail} \\
\midrule
Formula Calculation & 33.3 & 11.3 \\
Counting & 40.5 & 21.4 \\
Capitalization & 87.5 & 47.8 \\
\bottomrule
\end{tabular}
\end{subtable}
\label{tab:direct_prompting}
\end{table}

%% file: sections/analyses.tex
\section{Token-Level Memorization Analysis}
In this section, we use \framework\ to perform fine-grained token-level analysis. We begin by describing our experimental setup and how we select essential reasoning steps to analyze. We then apply \framework\ to compare token-level memorization patterns across tasks, distributions (base vs.\ long-tail), and correctness (correct vs.\ incorrect reasoning).

\input{figures/boxen_three_way}

\subsection{Experimental Setup and Step Selection}

Our analyses aim to contrast the model’s token-level memorization behavior between base and long-tail input distributions, as well as between correct and incorrect CoT reasoning chains.

In analyzing reasoning failures, we focus on the \textbf{first erroneous step} in each flawed chain rather than the entire output sequence. This design choice is motivated by the observation that subsequent reasoning steps often suffer from cascading effects caused by earlier mistakes, so such errors may not be directly attributable to memorization. By isolating the first incorrect step, we target the \textit{initial point of failure} where memorization-related influences are most likely to manifest. This approach also aligns with recent practice in process supervision~\citep{luo2024improve,lightman2023let}, where Process Reward Models (PRMs) are commonly trained to identify the earliest incorrect step.

For each task and for both the base and long-tail settings, we uniformly sample 200 examples where the model's final answer is incorrect (the \textbf{wrong} set) and 200 examples where the final answer is correct (the \textbf{correct} set). To identify the reasoning steps that lead to incorrect final answers, we apply VersaPRM~\citep{zeng2025versaprm}, a multi-domain Process Reward Model trained across 14 diverse domains. A step is classified as erroneous if its PRM score (ranging from 0 to 1) below an empirically determined threshold of 0.9. We exclude non-substantive statements like ``Let's verify'' or ``Let's think step by step''. We focus on the first one below the threshold as the first erroneous step for analysis. In practice, PRMs are not always reliable. They may miss reasoning errors and assign overly high scores even when the final answer is wrong. To ensure error in the selected step, we exclude such cases from our \textbf{wrong} set and retain only examples containing at least one step with a PRM score below 0.9. For each example in the \textit{correct} set, we select the reasoning step with the lowest PRM score to provide a point of contrast. All subsequent analyses are conducted on these selected reasoning steps. We verify the reliability of VersaPRM in Appendix~\ref{app:ablation_prm_lerg}.

\subsection{Analyzing Memorization Patterns by Task, Distribution, and Correctness}
\label{subsec:analysing memorization}

Now we illustrate the presence of memorization across tasks, distributions and correctness. While each token may be predominantly influenced by different sources, $\text{STIM}_{max}$, as the maximum of all three sources of memorization, represents the highest level of contextual influence at each token. Therefore, we present our observations on the distribution of $\text{STIM}_{max}$.

% \Figref{fig:memorization_effect_boxen} illustrates the distributions of average top-5 dominance score tokens across the four tasks, contrasting Base (blue) vs. Long-tail (brown) distribution and Correct (solid) vs. Wrong (transparent) predictions. Here are our main findings:

\paragraph{More complex reasoning tasks have higher memorization score.}
Across all four settings, Applied Math and Formula Calculation consistently show higher dominance scores compared to Capitalization and Counting (\Figref{fig:boxen_task}). This pattern aligns with the nature of the tasks: Applied Math and Formula Calculation involve more complex reasoning, requiring the model to iteratively synthesize partial solutions at each step. In contrast, Counting and Capitalization demand relatively simple, local decisions during intermediate steps, with synthesis needed only at the final stage.

\paragraph{Memorization scores are higher in long-tail settings.}
In 3 of 4 tasks, long-tail examples (brown) show consistently higher memorization scores than base examples (blue) (\Figref{fig:boxen_dist}). This may seem counterintuitive: base examples contain more frequent entities, so one might expect greater reliance on memorized content. Instead, the elevated scores suggest that when faced with unfamiliar inputs, the model often inappropriately falls back on spurious patterns memorized during pretraining, leading to more errors. Additional evidence is provided in~\Secref{sec:identify}.

The exception is \textit{Counting}, where scores are similar across settings. This likely stems from the task’s rigid reasoning format (e.g., ``The \textit{X}th element is \textit{Y}''). Despite longer lists in the long-tail setting, the model typically makes its first mistake around the 8th step in both cases. As a result, error patterns remain structurally similar, leaving little room for memorization differences to emerge.
\looseness=-1

\paragraph{Distribution Shift Reverses the Role of Memorization.}

We observe a shift in the relationship between memorization and correctness across distributions (\Figref{fig:boxen_correct}). In the base settings, correct reasoning steps exhibit higher memorization scores than incorrect ones, suggesting that memorized content aligns well with the input and supports accurate reasoning. This is especially evident in more complex tasks like Applied Math and Formula Calculation, where correct solutions may involve retrieving familiar mathematical equation patterns seen during pretraining. In contrast, in the long-tail setting , incorrect reasoning steps have higher memorization scores, indicating that the model often falls back on memorized but ill-fitting patterns when faced with less familiar inputs. This suggests that memorization, while helpful in familiar contexts, can hinder generalization under distribution shift -- especially when the model applies high-frequency completions to novel or rare scenarios. The reversal in score patterns highlights how the same underlying memorization behavior can contribute to success or failure depending on the alignment between the input and the model’s pretraining data.

\input{tables/dominance}

\input{tables/memo_hitrate}

\section{Detecting the Wrong Token using \framework}
\label{sec:identify}

With our framework for computing token-level memorization scores across multiple sources, we now explore a key application: identifying tokens that cause reasoning failures. We assess whether high memorization scores can pinpoint erroneous tokens in flawed reasoning steps, so as to demonstrate the framework’s value as a diagnostic tool for understanding how memorized content drives errors in long-form generation.

For each example that the model answers wrongly, we use \texttt{gpt4-o} to identify the wrong tokens in the erroneous reasoning step identified by VersaPRM described in~\Secref{subsec:stim}. 
The exact prompt used for \texttt{gpt4-o} is in Appendix~\ref{app:identify_gpt4}.

\paragraph{Dominant Source of Erroneous Memorization.}

We begin by identifying the \textbf{most influential type of memorization} (local, mid-range, or long-range) for each wrong token across tasks and distributional settings. This analysis reveals which memorization source most often drives the model’s reasoning in cases where it generates incorrect outputs. The percentages reported in~\Tabref{tab:dominant_memorization_source} reflect the share of all identified error tokens where a particular source has the highest memorization score.

Across all tasks and distributions, \textbf{local memorization} consistently emerges as the most frequent driver of erroneous token generation, accounting for as high as 67\% of error tokens. This suggests that models often fall for spurious short-range patterns even in tasks requiring structured reasoning.

While this heavy reliance on local cues is generally undesirable, since local context carries minimal task-relevant information, its distributional dynamics reveal an interesting trend. Under distribution shift from base to long-tail, \textbf{high-reasoning tasks} such as \textit{Applied Math }and \textit{Formula Calculation} show a \textbf{notable decrease} in the proportion of errors attributable to local memorization. In contrast, \textbf{low-reasoning tasks} like \textit{Counting} and \textit{Capitalization} show little to no such reduction.

This suggests that when facing unfamiliar, long-tail inputs, the model is forced to abandon brittle local heuristics in high-reasoning tasks and may instead attempt to engage more global or structured reasoning strategies. However, for simpler tasks where local patterns suffice even in the long-tail, the model continues to rely on them.

\paragraph{Correspondence between high memorization and token error.}
To further examine the alignment between high memorization and reasoning error, we also report \textbf{Precision@k} and \textbf{Recall@k} for $k \in [1,3]$, because we find that \texttt{gpt-4o} at most identifies 3 wrong tokens in a reasoning step.

Precision@k measures whether the top-$k$ tokens with highest memorization scores are wrong tokens. It is defined as:

$\mathcal{W}: \text{Set of wrong tokens} $

$\mathcal{S}: \text{Token set with top-k memorization score}$
\begin{equation}
    Precision@k=\frac{\sum_{\text{token} \in \mathcal{W}}\mathbb{I}(\text{token $\in \mathcal{S}$})}{min(|{\mathcal{W}}|, k)}
\end{equation}

The denominator handles cases where the number of wrong tokens is less than $k$, when we take the minimum of the two.
The final score is the average Precision@k across all evaluation examples in each task and each distribution.

Recall@k measures whether the wrong tokens appear within the top-$k$ tokens with highest memorization scores. It is defined as:

\begin{equation}
\small
Recall@k = 
\begin{cases}
1, & \text{if any of the wrong tokens exists in} \\
& \text{top-k highest memorization tokens} \\
0, & \text{otherwise}
\end{cases}
\end{equation}

The final score is the average Recall@k across all examples in each task and distribution.

In~\Tabref{tab:precision_recall} we report Precision and Recall @1-3 (1 meaning the highest memorization score token and 3 meaning the top 3 high memorization score token) of each task using \textbf{\framework$_{max}$}, as well as aggregated over all tasks. For Precision and Recall of individual components (\framework$_{local}$, \framework$_{long}$, \framework$_{mid}$) see Appendix~\ref{app:individual}.

\paragraph{Token-level memorization scores meaningfully correlate with erroneous tokens}
For all 4 tasks, Precision and Recall@1–3 scores are well above random chance (calculation described in Appendix~\ref{app:random_chance}), especially in more complex reasoning tasks like \textit{Applied Math} and \textit{Formula Calculation}. This indicates that tokens with higher memorization scores are indeed more likely to be error-inducing.
\looseness=-1

\paragraph{Complex tasks show higher Precision and Recall@k}
\textit{Applied Math} and \textit{Formula Calculation} consistently have higher precision and recall across all levels compared to \textit{Counting} and \textit{Capitalization}. This suggests that errors in complex reasoning tasks are more likely to be associated with memorization.
\looseness=-1

\paragraph{Computational Complexity of Wrong Token Identification with \framework}
The overall complexity of identifying wrong tokens using \framework\ is $O(mn)$, where $m$ denotes the number of salient tokens selected by the token saliency method, and $n$ is the number of alternative candidate tokens considered for each target token.

\paragraph{Precision and recall improve with higher \textit{k}.}
Both precision and recall increase consistently from $k=1$ to $k=3$ across tasks and levels, indicating that while the top memorized token is not always the erroneous one, the true error is often among the top three. This suggests that our method can serve as an effective first-pass filter, narrowing down the set of candidate wrong tokens that require further verification.

% From results in ~\Secref{sec:identify}, we still see at least 20\% of examples whose main error tokens do not result from memorization. We perform case analyses on these items.

% \todo{Illustrative examples on non-memorization, presented in a table}

%% file: figures/boxen_three_way.tex
% \begin{figure}
%     \centering
%     \includegraphics[width=\textwidth]{images/boxen_top5_max.pdf}
%     \caption{Boxen plot showing the distribution of average top-5 dominance score tokens across the four tasks, contrasting Head (blue) vs. Long-tail (brown) distributions and Correct (dark) vs. Wrong (light) predictions.}
%     \label{fig:memorization_effect_boxen}
% \end{figure}

 \begin{figure*}[htbp]
  \centering
  \begin{subfigure}[b]{0.3\textwidth}
  \centering
  \includegraphics[height=.65\linewidth]{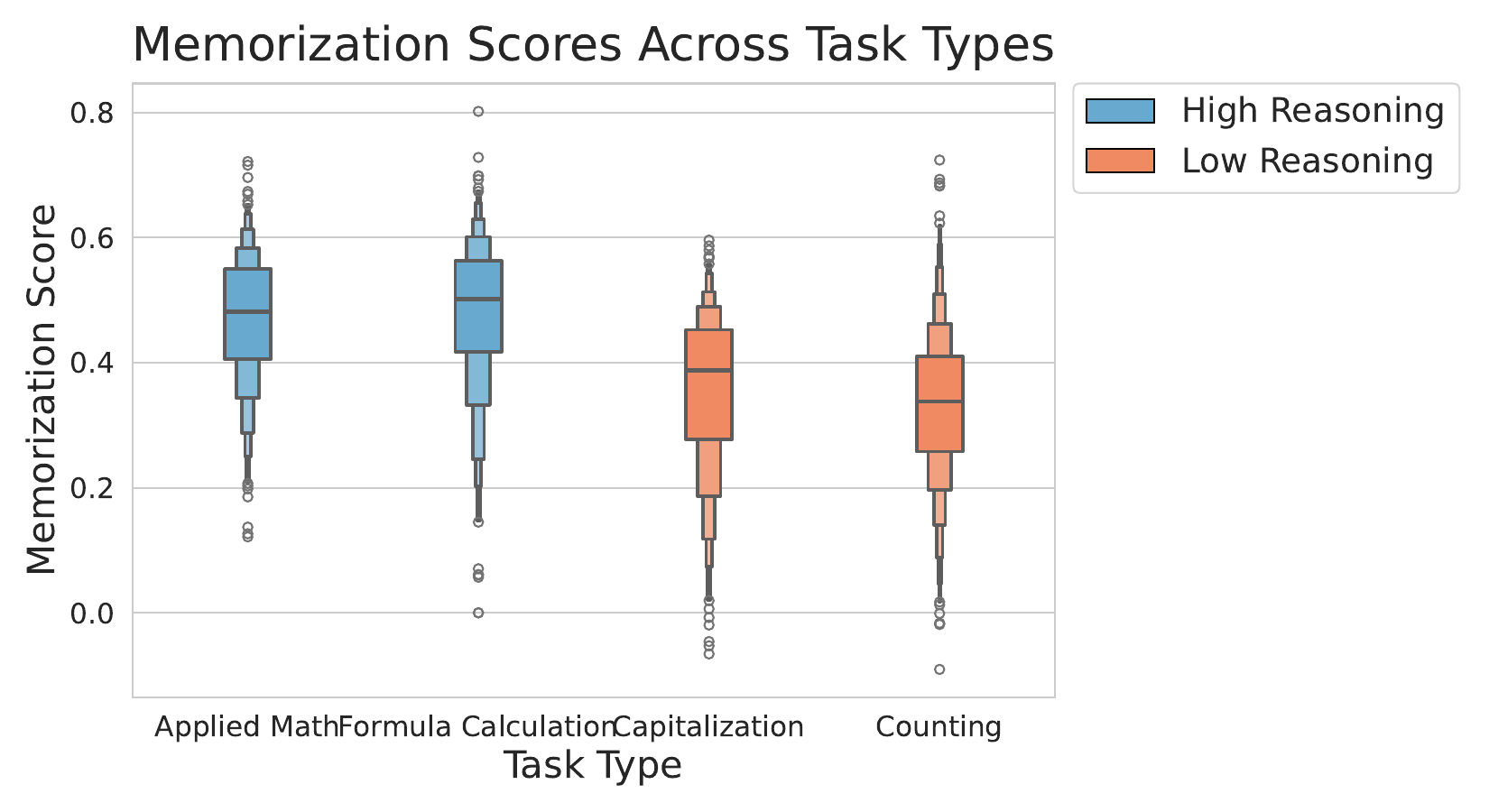}
  \caption{By Task}
  \label{fig:boxen_task}
  \end{subfigure}
  \hfill
  \begin{subfigure}[b]{0.3\textwidth}
  \centering
  \includegraphics[height=.65\linewidth]{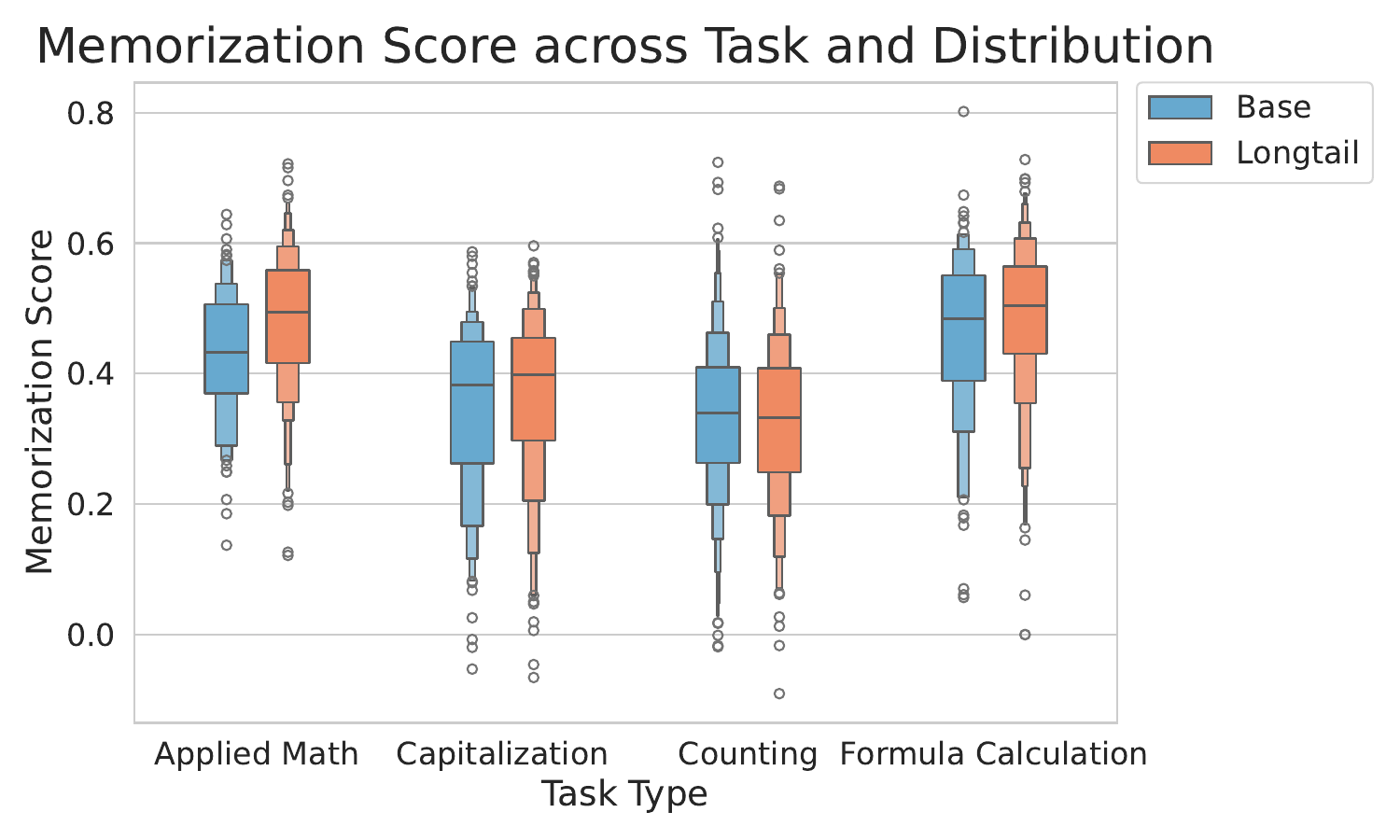}
  \caption{By Distribution}
  \label{fig:boxen_dist}
  \end{subfigure}%
  \hfill%
  \begin{subfigure}[b]{0.3\textwidth}
  \centering
  \includegraphics[height=.65\linewidth]{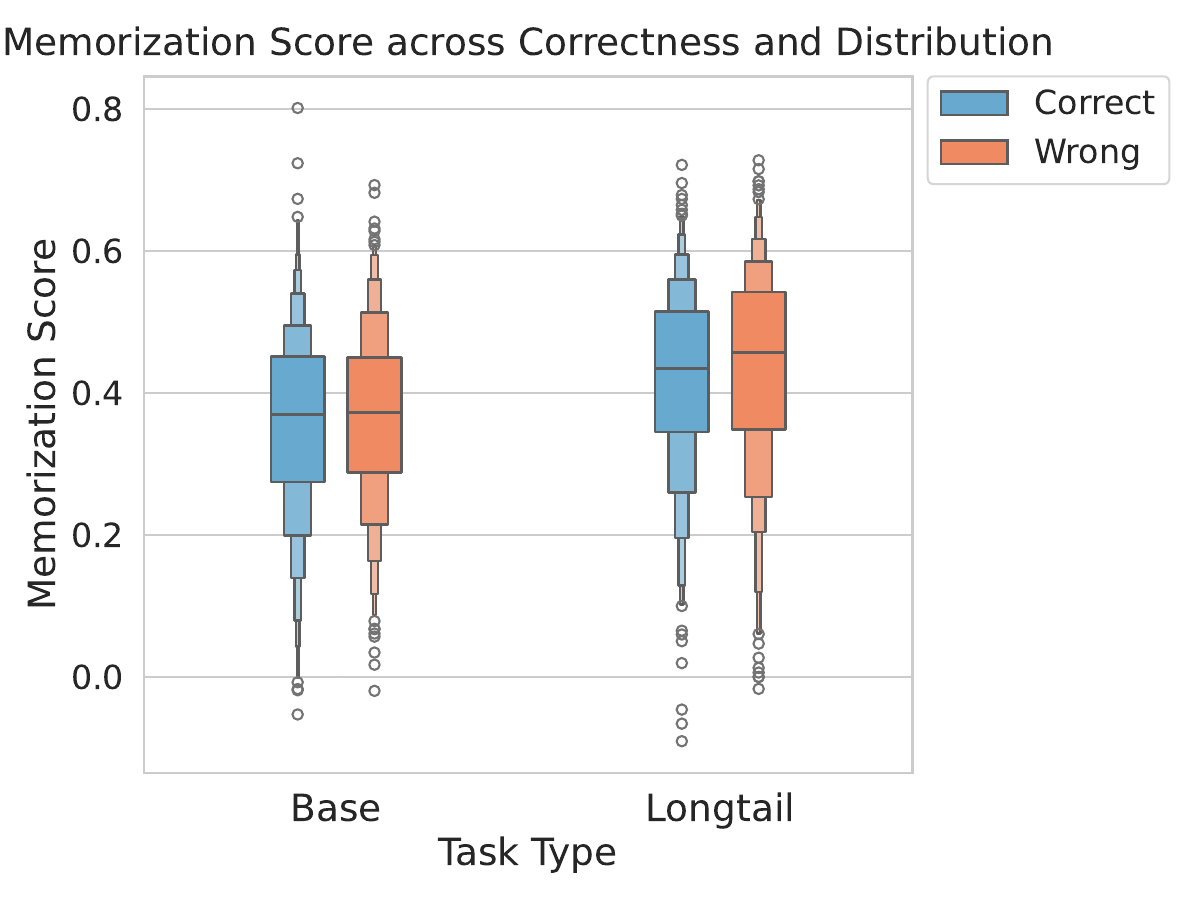}
  \caption{By Correctness}
  \label{fig:boxen_correct}
  \end{subfigure}
  \caption{Comparison of $\text{STIM}_{max}$ distribution between task reasoning complexity, distribution and correctness.
  \vspace{-1em}}
  \label{fig:three_figures}
 \end{figure*}

%% file: tables/dominance.tex
\begin{table}[hbt]
\centering
\caption{Dominant memorization source (\%) by task and distribution type. Each row shows the percentage of erroneous tokens where each source (Local, Mid, Long) was the most influential.}
\label{tab:dominant_memorization_source}

\small
\begin{tabular}{llccc}
\toprule
\textbf{Task} & \textbf{Dist.} & \textbf{Local} & \textbf{Mid} & \textbf{Long} \\
\midrule
\multirow{2}{*}{\makecell[l]{Applied \\ Math}}
& Base      & \textbf{0.673} & 0.127 & 0.200 \\
& Long-tail & \textbf{0.474} & 0.299 & 0.227 \\
\midrule
\multirow{2}{*}{\makecell[l]{Formula \\ Calculation}}
& Base      & \textbf{0.517} & 0.186 & 0.297 \\
& Long-tail & \textbf{0.443} & 0.253 & 0.304 \\
\midrule
\multirow{2}{*}{Counting}
& Base      & \textbf{0.520} & 0.208 & 0.272 \\
& Long-tail & \textbf{0.663} & 0.207 & 0.130 \\
\midrule
\multirow{2}{*}{Capitalization}
& Base      & \textbf{0.647} & 0.141 & 0.212 \\
& Long-tail & \textbf{0.649} & 0.185 & 0.166  \\
\bottomrule
\end{tabular}

\end{table}

%% file: tables/memo_hitrate.tex
\begin{table*}[t]
\centering
\caption{Precision and Recall @1–3 of $\text{STIM}_{\max}$ for identifying the wrong token in erroneous reasoning steps.
\vspace{-1em}}
\label{tab:precision_recall}
\begin{subtable}[t]{0.48\textwidth}
\centering
\caption{Precision@k}
\footnotesize
\begin{tabular}{llccc}
\toprule
\rowcolor{gray!25}
\textbf{Task} & \textbf{Level} & P@1 & P@2 & P@3 \\
\midrule
\multirow{2}{*}{\makecell[l]{Applied \\ Math}}
& $\text{STIM}_{\max}$ & 45.5 & 45.2 & 57.2 \\
& Random & 15.8 & 23.2 & 29.8 \\
\midrule
\multirow{2}{*}{\makecell[l]{Formula \\ Calc.}}
& $\text{STIM}_{\max}$ & 41.0 & 49.2 & 60.5 \\
& Random & 20.9 & 34.1 & 38.5 \\
\midrule
\multirow{2}{*}{Counting}
& $\text{STIM}_{\max}$ & 21.0 & 28.8 & 41.7 \\
& Random & 10.1 & 16.3 & 23.2 \\
\midrule
\multirow{2}{*}{Capitalization}
& $\text{STIM}_{\max}$ & 17.2 & 32.0 & 42.7 \\
& Random & 10.0 & 17.9 & 26.6 \\
\midrule
\multirow{2}{*}{All Tasks}
& $\text{STIM}_{\max}$ & \textbf{31.2} & \textbf{38.8} & \textbf{50.5} \\
& Random & 14.2 & 22.9 & 29.5 \\
\bottomrule
\end{tabular}
\end{subtable}
\hfill
\begin{subtable}[t]{0.48\textwidth}
\centering
\caption{Recall@k}
\footnotesize
\begin{tabular}{llccc}
\toprule
\rowcolor{gray!25}
\textbf{Task} & \textbf{Level} & R@1 & R@2 & R@3 \\
\midrule
\multirow{2}{*}{\makecell[l]{Applied \\ Math}}
& $\text{STIM}_{\max}$ & 40.5 & 55.0 & 67.0 \\
& Random & 12.8 & 24.8 & 35.2 \\
\midrule
\multirow{2}{*}{\makecell[l]{Formula \\ Calc.}}
& $\text{STIM}_{\max}$ & 37.5 & 54.0 & 68.0 \\
& Random & 15.5 & 29.2 & 40.7 \\
\midrule
\multirow{2}{*}{Counting}
& $\text{STIM}_{\max}$ & 19.5 & 31.0 & 45.5 \\
& Random & 13.0 & 25.5 & 37.7 \\
\midrule
\multirow{2}{*}{Capitalization}
& $\text{STIM}_{\max}$ & 17.5 & 34.5 & 46.0 \\
& Random & 11.4 & 27.4 & 40.1 \\
\midrule
\multirow{2}{*}{All Tasks}
& $\text{STIM}_{\max}$ & \textbf{28.8} & \textbf{43.6} & \textbf{56.6} \\
& Random & 13.2 & 26.7 & 38.4 \\
\bottomrule
\end{tabular}
\end{subtable}
\end{table*}

%% file: sections/disc_conclusion.tex
% \section{Discussion}
% different tasks can be measured by different metrics best
% % \subsection{What is beyond memorization?}
% % From results in ~\Secref{sec:identify}, we still see at least 20\% of examples whose main error tokens do not result from memorization. We perform case analyses on these items.

\section{Conclusion}
Our paper presented \framework, a novel diagnostic framework for fine-grained, token-level identification of memorization in \cotr\ reasoning, capturing multiple sources of memorization by analyzing both the input prompt and generated context. Our evaluation across diverse CoT tasks shows that memorization intensifies with task complexity and long-tail distribution shift, often leading to errors -- especially driven by local memorization.
Beyond aggregate trends, \framework\ reliably identifies error-inducing tokens, with high memorization scores correlating with incorrect tokens across tasks. This predictive signal underscores the practical value of token-level memorization as a lens on reasoning failures.
\framework\ offers a foundation for deeper insights into model behavior and toward developing more robust, genuinely reasoning-capable LLMs.

%% file: sections/limitations.tex
\section*{Limitations}
One limitation of our framework lies in the choice of token saliency method. We adopt LERG, a perturbation-based approach, to identify influential tokens in mid- and long-range memorization due to its favorable compute efficiency. However, LERG may miss finer-grained influence patterns compared to more computationally intensive alternatives, such as gradient-based methods or causal tracing approaches, which could yield more precise attributions.

Another constraint is the use of a pre-trained PRM (Process Reward Model) as the step verifier for detecting reasoning errors. While the verifier achieves strong performance in practice, it is not perfect and may occasionally mislabel erroneous or correct steps. Incorporating stronger verification models or ensemble-based approaches could further improve the robustness of our identification pipeline.

Finally, our analysis is limited by the availability of open-source language models with fully indexed pretraining corpora. Currently, only a few models, such as OLMo and Pythia meet these requirements, but Pythia is too weak to complete many tasks in \cotr\ reasoning. Widely used models either lack dataset transparency or are not fully open, restricting broader applicability of our framework across model scales in our work.

\section*{Risk}

\paragraph{Tracing of proprietary, sensitive information or PII.} \framework~might be used on mal-intentioned tasks, where researchers may prompt models produce sensitive information and then apply \framework\ to analyze which areas in the prompt contributes most for retrieving such information.

\paragraph{Environmental tax.} Another potential risk is increasing environmental burdens because we extensively ping infinigram API when searching through pretraining corpora, leading to extra usage of electricity and power.

\section*{Use and Distribution}

All data we collected through LLMs in our work are released publicly for usage and have been duly scrutinized by the authors. Our work does not collect information that can be used to identify individual people or contents that may be offensive.

Our framework \framework~may only be used for analysis and examinations of language models
followng the ethics guideline of the community. Using \framework~on mal-intentioned tasks is a potential threat, but the authors strongly condemn doing so.

%% file: sections/appendix.tex
% \section{Appendix}
% \label{sec:appendix}

\section{Head and Long-tail Data of each reasoning task}
\label{app:data_construction}
\input{tables/task_examples}

% for each task, cite source, tools we used, and each type of long-tail transformation

\paragraph{Applied Math}
The head data is the original test set of GSM8K \citep{cobbe2021training}. The Long-tail transformation process includes identifying the numerical entities in the original problems, shifting the numerical entities, and recalculating the final result. Following Varbench \citep{qian2024varbench}, we get the extracted numerical variables that are identified by LLMs and verified by experts. Then we apply digit-expansion, int-to-float-conversion, and base-changing transformation to those entities. For digit-expansion, we randomly sample two prime numbers $p_1$, $p_2$ between 10 to 30, and convert the number entity $var$ into $p_1\times var+p_2$; for int-to-float conversion, we divide the original variable by 100; for base-changing, we convert the numbers into base-2, limiting 8 bits of precision for floating numbers. Finally , we use the solution function in Varbench to get the final result with shifted value for each numeric variable.
\paragraph{Formula Calculation}
The head data is the formulas extracted from GSM8K solutions. We combine all the partial formulas in the step-by-step solutions into the final compositional formula. Similar to Applied Math long-tail transformation process, we change each numerical variable of the left expression into numbers with larger magnitude, floating numbers and base-2 number, keeping them be equal to the correspondent value in the applied math problem.
\paragraph{Counting}
We use \texttt{gpt4-o} to generate multiple fruits and then select the fruits with pretraining frequency having order of magnitude $10^3, 10^4, 10^5, 10^6, 10^7$. Then we randomly combine them to form the counting list, with the length ranging from $10, 20, 30, 40, 50$. Each list only contains two types of fruits.
\paragraph{Capitalization}
We first filter the book title datasets \citep{iwana2016judging} by restricting the length equal to $3, 5, 7, 9, 11$ and then randomly sample 300 examples for each length group. Finally, we lower the character in the titles to get the original string and only capitalize the first letter of the last word of the original string to acquire the long-tail entity.

Number of examples for each reasoning task and settings are shown in Table~\ref{tab:num_examples}.

\input{tables/num_examples}

\section{Identifying Wrong Tokens with \texttt{gpt4-o}}
\label{app:identify_gpt4}
We use \texttt{gpt4-o} to identify the wrong tokens in model's erroneous reasoning step for all tasks. The detailed prompts are shown in Table~\ref{tab:prompt_gpt4o}.
\input{tables/prompt_gpt4o}

\section{Calculating Random Chance Baseline for Precision@k and Recall@k}
\label{app:random_chance}
For precision@k, we calculate all the possible $k$ tokens' combinations from the wrong reasoning steps, which forms $C_n^k$ sets $\mathcal{S}_i$, where $n$ is number of candidate tokens. Then we calculate the proportion of wrong tokens in the $\mathcal{S}_i$ for each combination. Finally, we implement the average across all combinations:
\begin{equation}
    P@k_{random}=\frac{1}{C_n^k}\sum_{i=1}^{C_n^k}\frac{\sum_{\texttt{word} \in \mathcal{W}}\mathbb{I}( \ \texttt{token $\in \mathcal{S}_i$})}{\min(|\mathcal{W}|, k)}
\end{equation}

For recall@k, we calculate the proportion of randomly selected $k$ tokens in the wrong token set. Denote $n$ as the number of candidate tokens, $m$ as the number of wrong tokens, then $R@k_{random}$ is defined as:
\begin{equation}
    R@k_{random}=1-\frac{C_{n-m}^k}{C_n^k}
\end{equation}

\section{Reproducing Main Results on OLMo2 7B Instruct (\texttt{olmo-2-1124-7b-instruct})}
\label{app:reproducing}

Although our model choice is constrained by available open-source models with fully indexed pretraining corpora, we reproduce our analysis on OLMo2 7B Instruct, a model trained on the same data as OLMo2 13B Instruct but only smaller in size. We select one task with high reasoning complexity (Applied Math) and one task with low reasoning complexity (Capitalization). All generation and evaluation settings remain the same.

In summary, our additional experiments in OLMo2 7B Instruct show that all our findings and performance of STIM are \textbf{generalizable to new models}, at least of the same architecture and different size.

\subsection{Analyzing Memorization Patterns by Task, Distribution and Correctness (Section~\ref{subsec:analysing memorization}, Figure~\ref{fig:three_figures})}

Figure~\ref{fig:three_figures_7b} reproduces the original three findings of Section~\ref{subsec:analysing memorization}.

\paragraph{Original Finding 1: More complex reasoning tasks have higher memorization score} The mean of memorization score for Applied Math (high complexity) is 0.463 while the mean of memorization score for Capitalization (low complexity) is 0.38.

\paragraph{Original Finding 2: Memorization scores are higher in long-tail settings} The mean of memorization scores for base and long-tail tasks for Applied Math are 0.43 and 0.482; the mean of memorization scores for base and long-tail tasks for Capitalization are 0.346 and 0.415. For both tasks, the memorization scores are higher in long-tail settings.

\paragraph{Original Finding 3: Distribution Shift Reverses the Role of Memorization.}
The mean of memorization scores for correct and wrong examples in base tasks are 0.392 and 0.356, with correct examples higher; The mean of memorization scores for correct and wrong examples in long-tail tasks are 0.443 and 0.468, with wrong examples higher.

\input{figures/boxen_7b}

\subsection{Dominant Source of Erroneous Memorization (Section~\ref{sec:identify}, Table~\ref{tab:dominant_memorization_source})}

\Tabref{tab:dominant_memorization_source_7b} shows the dominant memorization source (\%) by task and distribution type. The two findings in the original papers both hold: 1) local memorization consistently emerges as the most frequent driver of erroneous token generation, although the actual percentage decreases (from 67\% to 57\%); this shows that smaller models may be less impacted by spurious short-range patterns than larger models 2) high reasoning task (Applied Math) show a notable decrease in the proportion of errors attributable to local memorization when shifting from base to long-tail distribution; low reasoning task (Capitalization) shows no reduction but rather some increase.

\input{tables/dominance_7b}

\subsection{Correspondence between high memorization and token error (Section~\ref{sec:identify}, Table~\ref{tab:precision_recall})}

\Tabref{tab:precision_recall_7b} shows the Precision@k and Recall@k performance for k=1,2,3.

\input{tables/memo_hitrate_7b}

\paragraph{Original Finding 1: Token-level memorization scores meaningfully correlate with erroneous tokens}
For both tasks, Precision and Recall @1-3 are well above random chance. Compared to the original result on OLMo2 13B Instruct, the Precision@k scores and most of Recall@k scores increase.

\paragraph{Original Finding 2: Complex tasks show higher Precision and Recall@k}
This is true from the results above across k=1,2,3.

\paragraph{Original Finding 3: Precision and recall improve with higher k}
This is also true for both Applied Math and Capitalization tasks.

\section{Ablation Studies on Process Reward Models and Token Saliency Models}
\label{app:ablation_prm_lerg}

To further verify the reliability of VersaPRM in identifying genuinely incorrect reasoning steps, we calculated the proportion of examples where GPT-4o did not detect any erroneous tokens within the steps selected by VersaPRM (as described in Section~\ref{sec:identify}). These proportions were 2.0\% for Applied Math and Formula Calculation, 0.5\% for Counting, and 2.5\% for Capitalization. Such low rates indicate that VersaPRM’s selection of incorrect steps is reasonably robust.

To assess the sensitivity of our wrong-token prediction results to the choice of token saliency method other than LERG, we conducted an ablation study using an alternative approach based on contrastive explanations~\citep{yin2022interpreting} on 20 sampled incorrect examples per task. Contrastive explanations are valuable for highlighting why a token was chosen over a specific alternative. However, CE introduces instability due to challenging foil selection and incurs higher computational costs for language generation. Overall, LERG offers more consistent and efficient attribution in this setting. For this particular ablation, we select foil tokens using the alternative 5 tokens with highest token probability, at the same decoding step of the target token.

In Table~\ref{tab:precision_recall_lerg} we report the Precision@k and Recall@k with each token saliency method.

For P@1, P@2, R@1, and R@2, STIM$_{max}$ (LERG) performs similarly to or better than STIM$_{max}$ (CE). For P@3 and R@3, the results are mixed, but the differences remain modest. These observations suggest that while the choice of token saliency method can influence STIM’s predictive performance, the impact is limited.
Finally, we would like to emphasize that the STIM framework is independent of the particular choice of PRM or token saliency method. Its overall predictive capacity could be further improved as more accurate PRMs for multi-domain reasoning and more advanced token-saliency techniques become available.

\input{tables/memo_hitrate_lerg_ablation}

\section{Individual Component Analysis: STIM$_{loc}$, STIM$_{mid}$ and STIM$_{long}$}
\label{app:individual}

Table~\ref{tab:precision_recall_indivdual} includes the predictive performance of each individual STIM score (local, mid, long) on wrong token identification and report their P@1 and R@1.

\input{tables/memo_hitrate_individual}

The results indicate that none of STIM$_{local}$, STIM$_{mid}$, or STIM$_{long}$ consistently achieve the highest P@1 or R@1 across all three sources. For Applied Math and Capitalization, STIM$_{local}$ shows the best performance, while for Formula Only and Counting, STIM$_{long}$ performs best. This variation in performance across individual scores suggests that STIM$_{max}$ is more robust overall. When aggregated across all tasks, STIM$_{max}$ achieves the highest P@1 and R@1.

\section{Prompt for Direct Answer and \cotr\ format of each reasoning task}
\label{app:prompt}

We use few-shot prompting methods for each task. The prompt formats in COT and Direct settings are shown in Table~\ref{tab:prompt_applied_base_cot} to Table~\ref{tab:prompt_cap_last_cot}

\input{tables/prompt_task}

%% file: tables/task_examples.tex
\begin{table*}[ht]
\centering
{\small
\begin{tabularx}{\textwidth}{>{\bfseries}l X X}
\hline
\rowcolor[gray]{0.9}
\textbf{Task} & \textbf{Base} & \textbf{Long-tail} \\
\hline

Applied Math &
\textbf{Question Prompt:} \newline James gets \textcolor{red}{10} new CDs.  Each CD cost \textcolor{red}{\$15}.  He gets them for \textcolor{red}{40\%} off.  He decides he doesn't like \textcolor{red}{5} of them and sells them for \textcolor{red}{40}. How much money was he out? \newline
\textcolor{mygray}{Decimal integers with small magnitude and Base-10 calculation} &
\textbf{Question Prompt:} \newline James gets \textcolor{red}{1010} new CDs. Each CD cost \textcolor{red}{\$1111}. He gets them for \textcolor{red}{101000\%} off. He decides he doesn't like \textcolor{red}{101} of them and sells them for \textcolor{red}{101000}. How much money was he out? \newline
\textcolor{mygray}{Decimal numbers with large magnitude; floating number; Base-2 calculation} \\

\hline

Formula Calculation &
\textbf{Question Prompt:} \newline What is \textcolor{red}{220} + \textcolor{red}{23} - \textcolor{red}{80} equal to? \newline
\textcolor{mygray}{Decimal integers with small magnitude and Base-10 calculation} &
\textbf{Question Prompt:} \newline What is \textcolor{red}{2437} + \textcolor{red}{270} - \textcolor{red}{897} equal to? \newline
\textcolor{mygray}{Decimal numbers with large magnitude; floating number; Base-2 calculation} \\

\hline

Counting &
\textbf{Question Prompt:} \newline Here is a list: \textcolor{red}{[apple, pear,$\cdots$, pear]}. How many times does `apple' appear on it? \newline
\textcolor{mygray}{Fruit's pretraining frequency $\geq 10^5$ and list length $\leq$ 20} &
\textbf{Question Prompt:} \newline Here is a list: \textcolor{red}{[keule, ugli,$\cdots$, keule]}. How many times does `ugli' appear on it? \newline
\textcolor{mygray}{Fruit's pretraining frequency $< 10^5$ or list length $>$ 20} \\

\hline

Capitalization &
\textbf{Question Prompt:} \newline Here is a string: \textcolor{red}{``history and obstinacy''}. Change the format of the string so that it can be a title. \newline
\textcolor{mygray}{Existing book titles and capitalize into title format} &
\textbf{Question Prompt:} \newline Here is a string: \textcolor{red}{``reasons to live''}. Change the format of the string so that only the first letter of the last word is capitalized. \newline
\textcolor{mygray}{Existing book titles and capitalize first letter of last word} \\

\hline
\end{tabularx}
}
\caption{Illustrative prompts contrasting standard (Base) and more complex (Long-tail) versions of reasoning tasks, with grey annotations explaining the difference in difficulty.}
\label{tab:head_lt_eg}
\end{table*}

%% file: tables/num_examples.tex
\begin{table*}[htbp] % table* for wide table across columns, [htbp] for placement
\centering
\caption{Number of Examples Per Task and Distributional Setting}
\label{tab:num_examples}
\small % Smaller font size for the table

\begin{tabular}{llc} % Two left-aligned columns (ll) and one centered column (c)
\toprule
\rowcolor{gray!25} % Optional: Highlight header row
\textbf{Task} & \textbf{Distribution} & \textbf{\# Examples} \\
\midrule

\multirow{3}{*}{\makecell[l]{Applied \\ Math}} % Multirow for 3 rows (1 Base + 2 Long-tail)
& Base (GSM8K)    & 1319 \\
& Long-tail (Digit Expansion)  & 1319 \\
& Long-tail (Integer to Float)  & 1319 \\
& Long-tail (Base 2)  & 1319 \\
\midrule

\multirow{3}{*}{\makecell[l]{Formula \\ Calculation}}
& Base (GSM8K)    & 1314 \\
& Long-tail (Digit Expansion)  & 1314 \\
& Long-tail (Integer to Float)  & 1314 \\
& Long-tail (Base 2)  & 1314 \\
\midrule

\multirow{3}{*}{Counting}
& Base (Length $\leq$ 20 \& frequent entities)      & 6000 \\
& Long-tail (Length $>$ 20 \& infrequent entities)  & 17500 \\
\midrule

\multirow{3}{*}{Capitalization}
& Base (Cap Title)        & 1500 \\
& Long-tail (Cap Last Word)  & 1500 \\
\bottomrule
\end{tabular}
\end{table*}

%% file: tables/prompt_gpt4o.tex
\begin{table*}[htbp]
    \centering
    \begin{tcolorbox}[
        colback=gray!3!white,
        colframe=gray!50!black,
        boxrule=0.5pt,
        width=13cm,
        arc=2mm,
        title=\text{Prompt Template for gpt4-o wrong token identification},
        fonttitle=\bfseries,
        coltitle=white,
    ]
    \textbf{System message:}
    
    \texttt{You are a helpful reasoning agent that can identify wrong tokens in another model's reasoning steps, by first generating your reasoning and then giving a final answer. You follow the examples given to you.}
    
    \textbf{User message}
    
    \texttt{Below you will see a reasoning question and a model's partial step-wise reasoning answer for it. The last step of the model's answer is wrong. Identify all the incorrect token(s) in that step that independently cause the step to be wrong, in "token, it's preceding token" format. Ignore the tokens whose error is caused by previous input tokens. You must give your answer from the given candidate tokens and choose at least one token. You will see two examples:}
    
    \texttt{Example 1}
    
    \texttt{Question: Last night Rick killed 1010 wolves and 1111 cougars while hunting. Today Rick killed 11 times as many wolves as cougars and 11 fewer cougars than the previous night. How many animals did Rick kill? (Calculate in Base-2)}
    
    \texttt{Step-wise Reasoning: Rick killed 11 * 1111 = 12221 wolves today.}
    
    \texttt{Candidate tokens: "Rick", preceded by ""; "killed", preceded by "Rick"; "11", preceded by "killed"; "*", preceded by "11"; "1111", preceded by "*"; "=", preceded by "1111"; "12221", preceded by "="; "wolves", preceded by "12221"; "today", preceded by "wolves"}
    
    \texttt{Reasoning: The token 1111 is semantically incorrect because at this step the model first needs to calculate the number of cougars which is 1111 - 11 = 1100, which is the correct number here. The token 12221 is also semantically incorrect. [...] Thus, 12221 is the incorrect token that directly causes the step to be wrong.}
    
    \texttt{Answer: "1111", preceded by "*"; "12221", preceded by "="}

    \vspace{0.5em}
    \textcolor{blue}{\texttt{[...one more example]}}
    \vspace{0.5em}
    
    \texttt{Now it's your turn: }
    
    \texttt{Question: Here is a list: [orange, banana, orange, banana, banana, banana, orange, orange, banana, orange]. How many times does 'orange' appear on it?} 
    
    \texttt{Step-wise Reasoning: To determine how many times the symbol 'orange' appears in the list [orange, banana, orange, banana, banana, banana, orange, orange, banana, orange], we can simply count each occurrence of 'orange'.}
    
    \texttt{The first element is 'orange'. }
    
    \texttt{The third element is 'orange'. }
    
    \texttt{The seventh element is 'orange'.}
    
    \texttt{The ninth element is 'orange'.}
    
    \texttt{Candidate tokens: “ninth”, preceded by “The”; “element”, preceded by “is”; “‘orange’”, preceded by ‘is’}
    
    \texttt{Reasoning:}

    \end{tcolorbox}
    \captionof{table}{Prompt of identifying wrong tokens for \texttt{gpt4-o}}
    \label{tab:prompt_gpt4o}
\end{table*}

%% file: figures/boxen_7b.tex
 \begin{figure*}[htbp]
  \centering
  \begin{subfigure}[b]{0.3\textwidth}
  \centering
  \includegraphics[height=.65\linewidth]{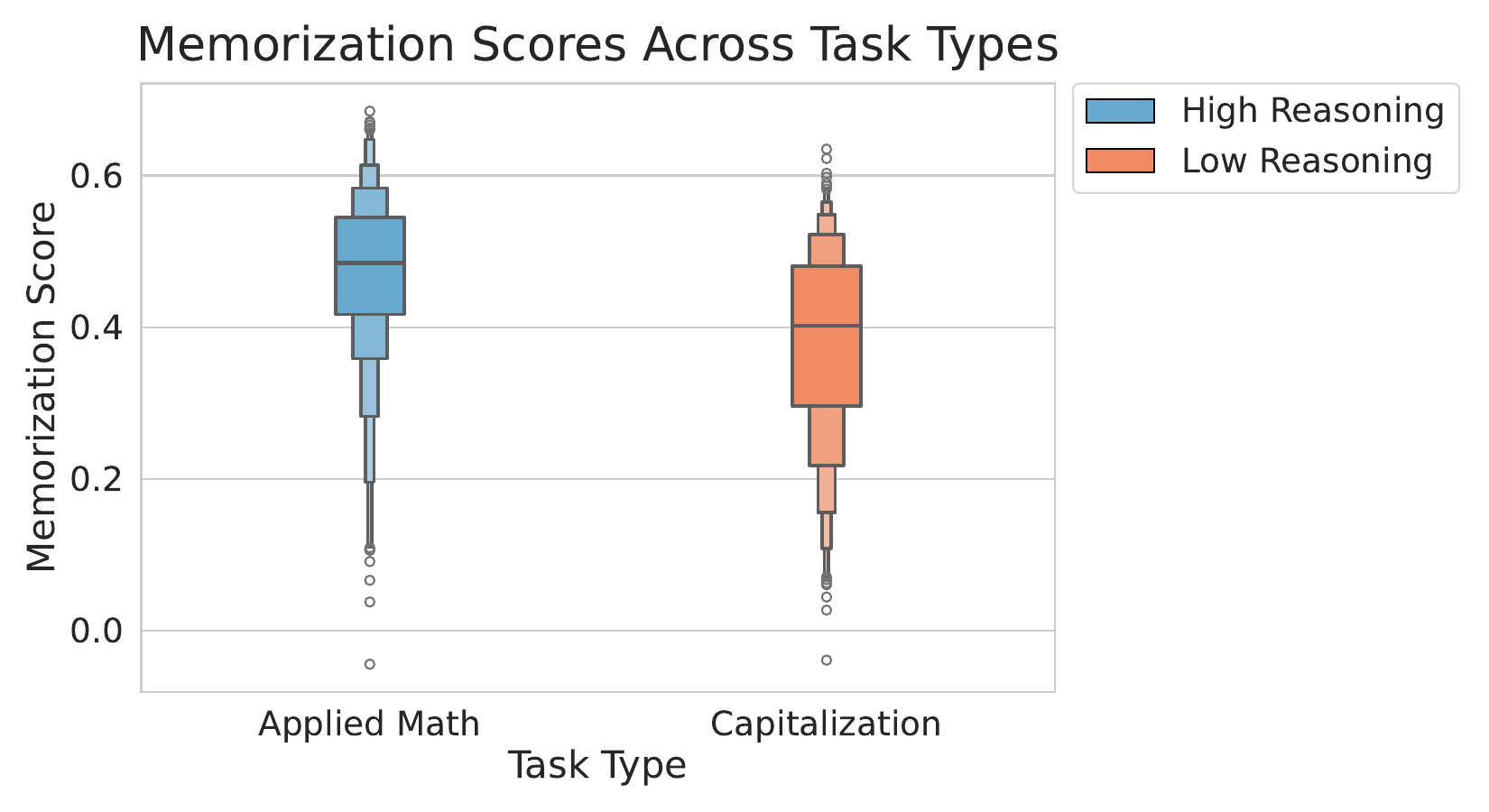}
  \caption{By Task}
  \label{fig:boxen_task}
  \end{subfigure}
  \hfill
  \begin{subfigure}[b]{0.3\textwidth}
  \centering
  \includegraphics[height=.65\linewidth]{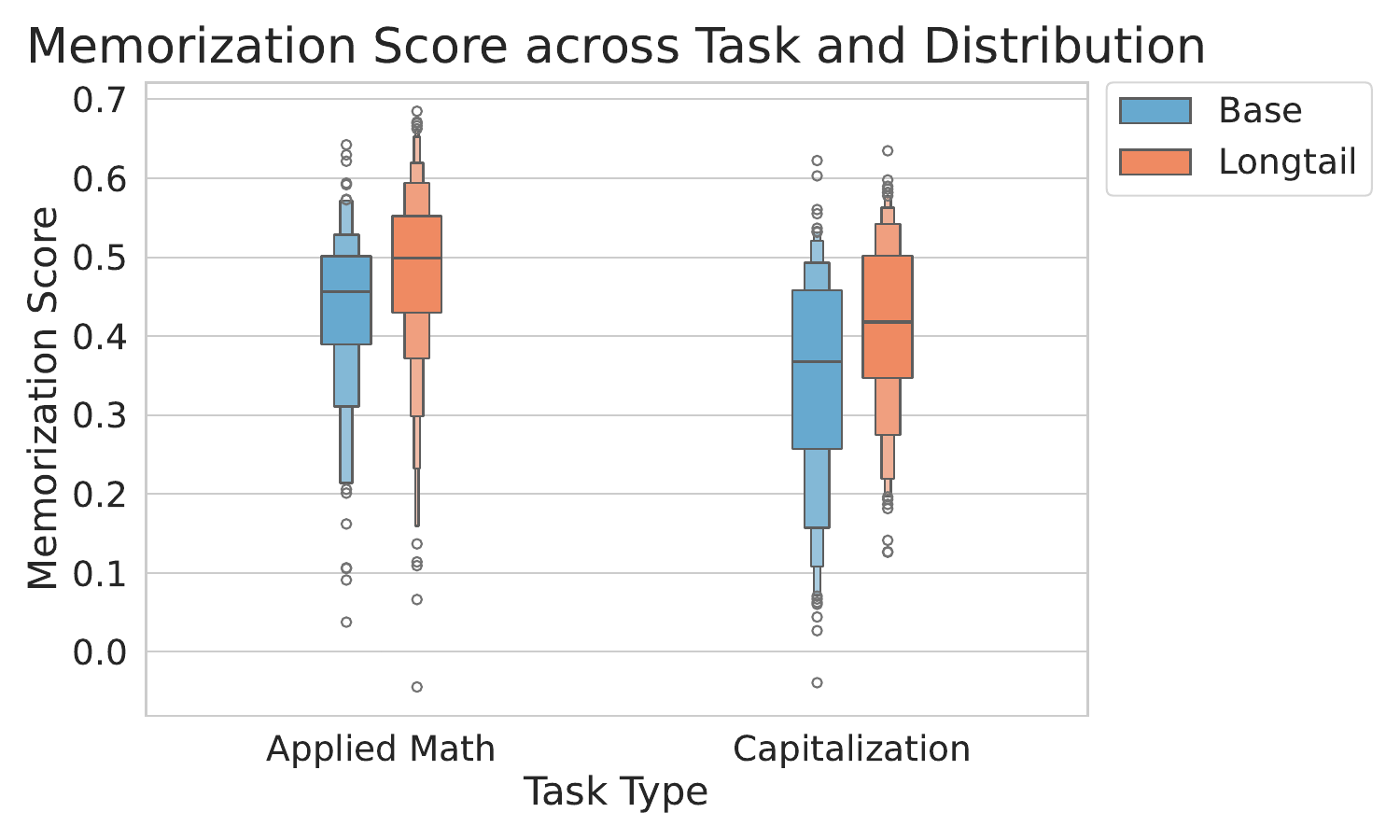}
  \caption{By Distribution}
  \label{fig:boxen_dist}
  \end{subfigure}%
  \hfill%
  \begin{subfigure}[b]{0.3\textwidth}
  \centering
  \includegraphics[height=.65\linewidth]{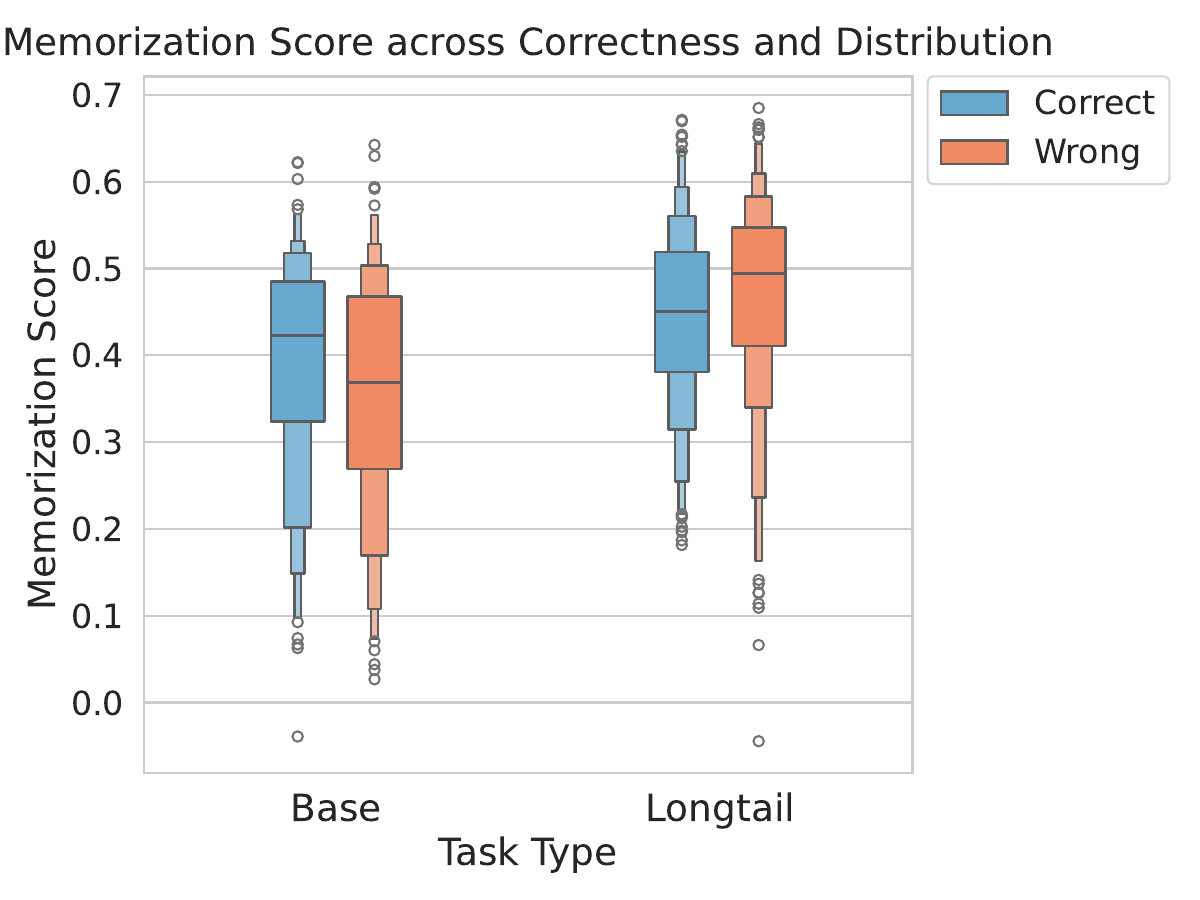}
  \caption{By Correctness}
  \label{fig:boxen_correct}
  \end{subfigure}
  \caption{Comparison of $\text{STIM}_{max}$ distribution between task reasoning complexity, distribution and correctness for OLMo2 7B Instruct.
  \vspace{-1em}}
  \label{fig:three_figures_7b}
 \end{figure*}

%% file: tables/dominance_7b.tex
\begin{table}[hbt]
\centering
\begin{tabular}{llccc}
\toprule
\textbf{Task} & \textbf{Dist.} & \textbf{\textbf{Local}} & \textbf{Mid} & \textbf{Long} \\
\midrule
\multirow{2}{*}{\makecell[l]{Applied \\ Math}} 
             & Base     & \textbf{0.571} & 0.242 & 0.187 \\
             & Longtail & \textbf{0.423} & 0.317 & 0.260 \\
\multirow{2}{*}{Capitalization} 
             & Base     & \textbf{0.476} & 0.169 & 0.355 \\
             & Longtail & \textbf{0.515} & 0.240 & 0.245 \\
\bottomrule
\end{tabular}
\caption{Dominant memorization source (\%) by task and distribution type for OLMo 7B Instruct. Each row shows the percentage of erroneous tokens where each source (Local, Mid, Long) was the most influential.}
\label{tab:dominant_memorization_source_7b}
\end{table}

%% file: tables/memo_hitrate_7b.tex
\begin{table*}[t]
\centering
\begin{subtable}[t]{0.48\textwidth}
\centering
\footnotesize
\begin{tabular}{llccc}
\toprule
\rowcolor{gray!25}
\textbf{Task} & \textbf{Level} & \textbf{1} & \textbf{2} & \textbf{3} \\
\midrule
\multirow{2}{*}{\makecell[l]{Applied \\ Math}} 
    & $\text{STIM}_{\max}$ & 0.460 & 0.513 & 0.574 \\
    & Random               & 0.173 & 0.279 & 0.317 \\
\multirow{2}{*}{Capitalization} 
    & $\text{STIM}_{\max}$ & 0.225 & 0.318 & 0.465 \\
    & Random               & 0.089 & 0.154 & 0.229 \\
\bottomrule
\end{tabular}
\caption{Precision@k}
\end{subtable}
\begin{subtable}[t]{0.48\textwidth}
\centering
\footnotesize
\begin{tabular}{llccc}
\toprule
\rowcolor{gray!25}
\textbf{Task} & \textbf{Level} & \textbf{1} & \textbf{2} & \textbf{3} \\
\midrule
\multirow{2}{*}{\makecell[l]{Applied \\ Math}} 
    & $\text{STIM}_{\max}$ & 0.385 & 0.580 & 0.645 \\
    & Random               & 0.140 & 0.274 & 0.381 \\
\multirow{2}{*}{Capitalization} 
    & $\text{STIM}_{\max}$ & 0.215 & 0.340 & 0.505 \\
    & Random               & 0.105 & 0.208 & 0.309 \\
\bottomrule
\end{tabular}
\caption{Recall@k}
\end{subtable}
\caption{Precision and Recall @1–3 of $\text{STIM}_{\max}$ for identifying the wrong token in erroneous reasoning steps for OLMo2 7B Instruct.}
\label{tab:precision_recall_7b}
\end{table*}

%% file: tables/memo_hitrate_lerg_ablation.tex
\begin{table*}[ht]
\centering
\begin{subtable}[t]{0.4\textwidth}
\centering
\footnotesize
\begin{tabular}{llccc}
\toprule
\rowcolor{gray!25}
\textbf{Task} & \textbf{Method} & \textbf{P@1} & \textbf{P@2} & \textbf{P@3} \\
\midrule
\multirow{3}{*}{\makecell[l]{Applied \\ Math}} 
  & $\text{STIM}_{\max}$ (CE)   & 0.400 & 0.450 & 0.583 \\
  & $\text{STIM}_{\max}$ (LERG) & 0.400 & 0.450 & 0.575 \\
  & Random                      & 0.225 & 0.318 & 0.317 \\
\multirow{3}{*}{Formula Only} 
  & $\text{STIM}_{\max}$ (CE)   & 0.350 & 0.475 & 0.533 \\
  & $\text{STIM}_{\max}$ (LERG) & 0.400 & 0.500 & 0.533 \\
  & Random                      & 0.256 & 0.442 & 0.445 \\
\multirow{3}{*}{Counting} 
  & $\text{STIM}_{\max}$ (CE)   & 0.100 & 0.325 & 0.325 \\
  & $\text{STIM}_{\max}$ (LERG) & 0.100 & 0.325 & 0.475 \\
  & Random                      & 0.082 & 0.156 & 0.235 \\
\multirow{3}{*}{Capitalization} 
  & $\text{STIM}_{\max}$ (CE)   & 0.200 & 0.300 & 0.500 \\
  & $\text{STIM}_{\max}$ (LERG) & 0.250 & 0.350 & 0.450 \\
  & Random                      & 0.096 & 0.179 & 0.268 \\
\bottomrule
\end{tabular}
\caption{P@k}
\end{subtable}
\hspace{5em}
\begin{subtable}[t]{0.4\textwidth}
\centering
\footnotesize
\begin{tabular}{llccc}
\toprule
\rowcolor{gray!25}
\textbf{Task} & \textbf{Method} & \textbf{R@1} & \textbf{R@2} & \textbf{R@3} \\
\midrule
\multirow{3}{*}{\makecell[l]{Applied \\ Math}} 
  & $\text{STIM}_{\max}$ (CE)   & 0.350 & 0.550 & 0.700 \\
  & $\text{STIM}_{\max}$ (LERG) & 0.350 & 0.550 & 0.700 \\
  & Random                      & 0.159 & 0.301 & 0.417 \\
\multirow{3}{*}{Formula Only} 
  & $\text{STIM}_{\max}$ (CE)   & 0.350 & 0.500 & 0.550 \\
  & $\text{STIM}_{\max}$ (LERG) & 0.400 & 0.550 & 0.600 \\
  & Random                      & 0.211 & 0.369 & 0.500 \\
\multirow{3}{*}{Counting} 
  & $\text{STIM}_{\max}$ (CE)   & 0.100 & 0.350 & 0.350 \\
  & $\text{STIM}_{\max}$ (LERG) & 0.100 & 0.350 & 0.500 \\
  & Random                      & 0.082 & 0.163 & 0.244 \\
\multirow{3}{*}{Capitalization} 
  & $\text{STIM}_{\max}$ (CE)   & 0.200 & 0.300 & 0.500 \\
  & $\text{STIM}_{\max}$ (LERG) & 0.250 & 0.350 & 0.450 \\
  & Random                      & 0.121 & 0.241 & 0.357 \\
\bottomrule
\end{tabular}
\caption{R@k}
\end{subtable}
\caption{Precision (P@k) and Recall (R@k) at rank $k$ across tasks and methods, ablating on CE and LERG.}
\label{tab:precision_recall_lerg}
\end{table*}

%% file: tables/memo_hitrate_individual.tex
\begin{table*}[ht]
\centering
\begin{subtable}[t]{0.48\textwidth}
\centering
\footnotesize
\begin{tabular}{llc}
\toprule
\rowcolor{gray!25}
\textbf{Task} & \textbf{Method} & \textbf{P@1} \\
\midrule
\multirow{5}{*}{\makecell[l]{Applied \\ Math}} 
  & Local       & 0.440 \\
  & Mid         & 0.355 \\
  & Long        & 0.410 \\
  & $\text{STIM}_{\max}$ & 0.455 \\
  & Random      & 0.158 \\
\multirow{5}{*}{Formula Only} 
  & Local       & 0.375 \\
  & Mid         & 0.400 \\
  & Long        & 0.435 \\
  & $\text{STIM}_{\max}$ & 0.410 \\
  & Random      & 0.209 \\
\multirow{5}{*}{Counting} 
  & Local       & 0.195 \\
  & Mid         & 0.240 \\
  & Long        & 0.240 \\
  & $\text{STIM}_{\max}$ & 0.215 \\
  & Random      & 0.103 \\
\multirow{5}{*}{Capitalization} 
  & Local       & 0.180 \\
  & Mid         & 0.155 \\
  & Long        & 0.150 \\
  & $\text{STIM}_{\max}$ & 0.175 \\
  & Random      & 0.100 \\
\multirow{5}{*}{All Tasks} 
  & Local       & 0.298 \\
  & Mid         & 0.288 \\
  & Long        & 0.309 \\
  & $\text{STIM}_{\max}$ & 0.314 \\
  & Random      & 0.143 \\
\bottomrule
\end{tabular}
\caption{P@1}
\end{subtable}
\begin{subtable}[t]{0.48\textwidth}
\centering
\footnotesize
\begin{tabular}{llc}
\toprule
\rowcolor{gray!25}
\textbf{Task} & \textbf{Method} & \textbf{R@1} \\
\midrule
\multirow{5}{*}{\makecell[l]{Applied \\ Math}} 
  & Local       & 0.400 \\
  & Mid         & 0.305 \\
  & Long        & 0.350 \\
  & $\text{STIM}_{\max}$ & 0.405 \\
  & Random      & 0.129 \\
\multirow{5}{*}{Formula Only} 
  & Local       & 0.345 \\
  & Mid         & 0.375 \\
  & Long        & 0.400 \\
  & $\text{STIM}_{\max}$ & 0.375 \\
  & Random      & 0.155 \\
\multirow{5}{*}{Counting} 
  & Local       & 0.185 \\
  & Mid         & 0.220 \\
  & Long        & 0.220 \\
  & $\text{STIM}_{\max}$ & 0.200 \\
  & Random      & 0.130 \\
\multirow{5}{*}{Capitalization} 
  & Local       & 0.180 \\
  & Mid         & 0.155 \\
  & Long        & 0.145 \\
  & $\text{STIM}_{\max}$ & 0.175 \\
  & Random      & 0.110 \\
\multirow{5}{*}{All Tasks} 
  & Local       & 0.278 \\
  & Mid         & 0.264 \\
  & Long        & 0.279 \\
  & $\text{STIM}_{\max}$ & 0.289 \\
  & Random      & 0.131 \\
\bottomrule
\end{tabular}
\caption{R@1}
\end{subtable}
\caption{Precision (P@1) and Recall (R@1) across tasks and methods.}
\label{tab:precision_recall_indivdual}
\end{table*}

%% file: tables/prompt_task.tex
\begin{table*}[htbp]
    \centering
    \begin{tcolorbox}[
        colback=gray!3!white,
        colframe=gray!50!black,
        boxrule=0.5pt,
        width=15cm,
        arc=2mm,
        title=\text{Prompt Template for Applied Math, Base, CoT},
        fonttitle=\bfseries,
        coltitle=white,
    ]
    \texttt{Instruction: Answer the given question. You will end your response with a sentence in the format of `So the answer is <number>.'}
    
    \texttt{Question: There are 15 trees in the grove. Grove workers will plant trees in the grove today. After they are done, there will be 21 trees. How many trees did the grove workers plant today?}
    
    \texttt{Answer: There are 15 trees originally. Then there were 21 trees after some more were planted. So there must have been 21 - 15 = 6. So the answer is 6.}
    
    \vspace{0.5em}
    \textcolor{blue}{\texttt{[...7 more examples]}}
    \vspace{0.5em}
    
    \texttt{Instruction: Answer the given question. You will end your response with a sentence in the format of `So the answer is <number>.'}
    
    \texttt{Question: John drives for 3 hours at a speed of 60 mph and then turns around because he realizes he forgot something very important at home. He tries to get home in 4 hours but spends the first 2 hours in standstill traffic. He spends the next half-hour driving at a speed of 30mph, before being able to drive the remaining time of the 4 hours going at 80 mph. How far is he from home at the end of those 4 hours?}
    
    \texttt{Answer:}
    \end{tcolorbox}
    \captionof{table}{Prompt of calculating math word problems (CoT setting)}
    \label{tab:prompt_applied_base_cot}
\end{table*}

\begin{table*}[htbp]
    \centering
    \begin{tcolorbox}[
        colback=gray!3!white,
        colframe=gray!50!black,
        boxrule=0.5pt,
        width=15cm,
        arc=2mm,
        title=\text{Prompt Template for Applied Math, Longtail, CoT},
        fonttitle=\bfseries,
        coltitle=white,
    ]
    \texttt{Instruction: Assuming that all numbers are in base-2 where the digits are "01". Answer the given question. You will end your response with a sentence in the format of `So the answer is <number>.'}
    
    \texttt{Question: There are 1111 trees in the grove. Grove workers will plant trees in the grove today. After they are done, there will be 10101 trees. How many trees did the grove workers plant today?}
    
    \texttt{Answer: There are 1111 trees originally. Then there were 10101 trees after some more were planted. So there must have been 10101 - 1111 = 110. So the answer is 110.}
    
    \vspace{0.5em}
    \textcolor{blue}{\texttt{[...7 more examples]}}
    \vspace{0.5em}
    
    \texttt{Instruction: Assuming that all numbers are in base-2 where the digits are "01". Answer the given question. You will end your response with a sentence in the format of `So the answer is <number>.'}
    
    \texttt{Question: John drives for 11 hours at a speed of 111100 mph and then turns around because he realizes he forgot something very important at home. He tries to get home in 100 hours but spends the first 10 hours in standstill traffic. He spends the next half-hour driving at a speed of 11110mph, before being able to drive the remaining time of the 100 hours going at 1010000 mph. How far is he from home at the end of those 100 hours?}
    
    \texttt{Answer:}
    \end{tcolorbox}
    \captionof{table}{Prompt of implementing base-2 calculation in math word problems (CoT setting)}
    \label{tab:prompt_applied_longtail_cot}
\end{table*}

\begin{table*}[htbp]
    \centering
    \begin{tcolorbox}[
        colback=gray!3!white,
        colframe=gray!50!black,
        boxrule=0.5pt,
        width=13cm,
        arc=2mm,
        title=\text{Prompt Template for Formula Calculation, Base, Direct},
        fonttitle=\bfseries,
        coltitle=white,
    ]
    \texttt{Instruction: Answer the given question. You will end your response with a sentence in the format of `So the answer is <number>.'}
    
    \texttt{Question: What is 32 + 42 - 35 equal to?}
    
    \texttt{Answer: So the answer is 39.}
    
    \vspace{0.5em}
    \textcolor{blue}{\texttt{[...7 more examples]}}
    \vspace{0.5em}
    
    \texttt{Instruction: Answer the given question. You will end your response with a sentence in the format of `So the answer is <number>.'}
    
    \texttt{Question: What is (16 - 3 - 4) * 2 equal to?}
    
    \texttt{Answer:}
    \end{tcolorbox}
    \captionof{table}{Prompt of formula calculation for base-10 (Direct setting)}
    \label{tab:prompt_formula_base_direct}
\end{table*}

\begin{table*}[htbp]
    \centering
    \begin{tcolorbox}[
        colback=gray!3!white,
        colframe=gray!50!black,
        boxrule=0.5pt,
        width=15cm,  % Increased width for longer binary expressions
        arc=2mm,
        title=\text{Prompt Template for Formula Calculation, Longtail, Direct},
        fonttitle=\bfseries,
        coltitle=white,
    ]
    \texttt{Instruction: Assuming that all numbers are in base-2 where the digits are "01". Answer the given question. You will end your response with a sentence in the format of `So the answer is <number>.'}
    
    \texttt{Question: What is 100000 + 101010 - 100011 equal to?}
    
    \texttt{Answer: So the answer is 100111.}
    
    \vspace{0.5em}
    \textcolor{blue}{\texttt{[...7 more examples]}}
    \vspace{0.5em}
    
    \texttt{Instruction: Assuming that all numbers are in base-2 where the digits are "01". Answer the given question. You will end your response with a sentence in the format of `So the answer is <number>.'}
    
    \texttt{Question: What is (10010110 - 111100 / 1100100 * 10010110) * 110 equal to?}
    
    \texttt{Answer:}
    \end{tcolorbox}
    \captionof{table}{Prompt of formula calculation in base-2 (Direct setting)}
    \label{tab:prompt_formula_longtail_direct}
\end{table*}

\begin{table*}[htbp]
    \centering
    \begin{tcolorbox}[
        colback=gray!3!white,
        colframe=gray!50!black,
        boxrule=0.5pt,
        width=13cm,
        arc=2mm,
        title=\text{Prompt Template for Formula Calculation, Base, CoT},
        fonttitle=\bfseries,
        coltitle=white,
    ]
    \texttt{Instruction: Answer the given question. You will end your response with a sentence in the format of `So the answer is <number>.'}
    
    \texttt{Question: What is 32 + 42 - 35 equal to?}
    
    \texttt{Answer: To calculate 32 + 42 - 35, we need to first calculate 32 + 42. 32 + 42 = 74. Then we need to calculate 74 - 35. 74 - 35 = 39. So the answer is 39.}
    
    \vspace{0.5em}
    \textcolor{blue}{\texttt{[...7 more examples]}}
    \vspace{0.5em}
    
    \texttt{Instruction: Answer the given question. You will end your response with a sentence in the format of `So the answer is <number>.'}
    
    \texttt{Question: What is (16 - 3 - 4) * 2 equal to?}
    
    \texttt{Answer:}
    \end{tcolorbox}
    \captionof{table}{Prompt of formula calculation in base-10 (CoT setting)}
    \label{tab:prompt_formula_base_cot}
\end{table*}

\begin{table*}[htbp]
    \centering
    \begin{tcolorbox}[
        colback=gray!3!white,
        colframe=gray!50!black,
        boxrule=0.5pt,
        width=15cm,
        arc=2mm,
        title=\text{Prompt Template for Formula Calculation, Longtail, CoT},
        fonttitle=\bfseries,
        coltitle=white,
    ]
    \texttt{Instruction: Assuming that all numbers are in base-2 where the digits are "01". Answer the given question. You will end your response with a sentence in the format of `So the answer is <number>.'}
    
    \texttt{Question: What is 100000 + 101010 - 100011 equal to?}
    
    \texttt{Answer: To calculate 100000 + 101010 - 100011, we need to first calculate 100000 + 101010. 100000 + 101010 = 1001010. Then we need to calculate 1001010 - 100011. 1001010 - 100011 = 100111. So the answer is 100111.}
    
    \vspace{0.5em}
    \textcolor{blue}{\texttt{[...7 more examples]}}
    \vspace{0.5em}
    
    \texttt{Instruction: Assuming that all numbers are in base-2 where the digits are "01". Answer the given question. You will end your response with a sentence in the format of `So the answer is <number>.'}
    
    \texttt{Question: What is (10010110 - 111100 / 1100100 * 10010110) * 110 equal to?}
    
    \texttt{Answer:}
    \end{tcolorbox}
    \captionof{table}{Prompt of formula calculation in base-2 (CoT setting)}
    \label{tab:prompt_formula_longtail_cot}
\end{table*}

\begin{table*}[htbp]
    \centering
    \begin{tcolorbox}[
        colback=gray!3!white,
        colframe=gray!50!black,
        boxrule=0.5pt,
        width=13cm,
        arc=2mm,
        title=\text{Prompt Template for Counting, Base, Direct},
        fonttitle=\bfseries,
        coltitle=white,
    ]
    \texttt{Instruction: Answer the given question. You will end your response with a sentence in the format of 'So the answer is <number>.'}
    
    \texttt{Question: Here is a list: [orange, orange, orange, orange, orange, orange, orange, apple, orange, orange]. How many times does 'apple' appear on it?}
    
    \texttt{Answer: So the answer is 1.}
    
    \vspace{0.5em}
    \textcolor{blue}{\texttt{[...7 more examples]}}
    \vspace{0.5em}
    
    \texttt{Instruction: Answer the given question. You will end your response with a sentence in the format of 'So the answer is <number>.'}
    
    \texttt{Question: Here is a list: [pear, pear, pear, pear, pear, pear, pear, apple, pear, pear]. How many times does 'apple' appear on it?}
    
    \texttt{Answer:}
    \end{tcolorbox}
    \captionof{table}{Prompt of counting common fruits (Direct setting)}
    \label{tab:prompt_counting_base_direct}
\end{table*}

\begin{table*}[htbp]
    \centering
    \begin{tcolorbox}[
        colback=gray!3!white,
        colframe=gray!50!black,
        boxrule=0.5pt,
        width=13cm,
        arc=2mm,
        title=\text{Prompt Template for Counting, Longtail, Direct},
        fonttitle=\bfseries,
        coltitle=white,
    ]
    \texttt{Instruction: Answer the given question. You will end your response with a sentence in the format of 'So the answer is <number>.'}
    
    \texttt{Question: Here is a list: [pitanga, pitanga, pitanga, pitanga, pitanga, pitanga, pitanga, ugli, pitanga, pitanga]. How many times does 'ugli' appear on it?}
    
    \texttt{Answer: So the answer is 1.}
    
    \vspace{0.5em}
    \textcolor{blue}{\texttt{[...7 more examples]}}
    \vspace{0.5em}
    
    \texttt{Instruction: Answer the given question. You will end your response with a sentence in the format of 'So the answer is <number>.'}
    
    \texttt{Question: Here is a list: [ugli, ugli, ugli, pitanga, pitanga, pitanga, ugli, ugli, pitanga, pitanga]. How many times does 'ugli' appear on it?}
    
    \texttt{Answer:}
    \end{tcolorbox}
    \captionof{table}{Prompt of counting uncommon fruits (Direct setting)}
    \label{tab:prompt_counting_longtail_direct}
\end{table*}

\begin{table*}[htbp]
    \centering
    \begin{tcolorbox}[
        colback=gray!3!white,
        colframe=gray!50!black,
        boxrule=0.5pt,
        width=15cm,
        arc=2mm,
        title=\text{Prompt for Counting, Base, CoT},
        fonttitle=\bfseries,
        coltitle=white,
    ]
    \texttt{Instruction: Answer the given question. You will end your response with a sentence in the format of 'So the answer is <number>.'}
    
    \texttt{Question: Here is a list: [orange, orange, orange, orange, orange, orange, orange, apple, orange, orange]. How many times does 'apple' appear on it?}
    
    \texttt{Answer: Let's think step by step. To determine how many times the symbol 'apple' appears in the list [orange, orange, orange, orange, orange, orange, orange, apple, orange, orange], we can simply count the occurrences of 'apple' within the list.}
    \texttt{Looking at the list, we see:}
    \texttt{- There are eight 'orange' symbols.}
    \texttt{- There is one 'apple' symbol.}
    \texttt{So, 'apple' appears once in the list.}
    \texttt{So the answer is 1.}
    
    \vspace{0.5em}
    \textcolor{blue}{\texttt{[...7 more examples]}}
    \vspace{0.5em}
    
    \texttt{Instruction: Answer the given question. You will end your response with a sentence in the format of 'So the answer is <number>.'}
    
    \texttt{Question: Here is a list: [apple, apple, apple, orange, orange, orange, orange, apple, orange, orange]. How many times does 'apple' appear on it?}
    
    \texttt{Answer:}
    \end{tcolorbox}
    \captionof{table}{Prompt of counting common fruits (CoT setting)}
    \label{tab:prompt_counting_base_CoT}
\end{table*}

\begin{table*}[htbp]
    \centering
    \begin{tcolorbox}[
        colback=gray!3!white,
        colframe=gray!50!black,
        boxrule=0.5pt,
        width=15cm,
        arc=2mm,
        title=\text{Prompt Template for Counting, Longtail, COT},
        fonttitle=\bfseries,
        coltitle=white,
    ]
    \texttt{Instruction: Answer the given question. You will end your response with a sentence in the format of 'So the answer is <number>.'}
    
    \texttt{Question: Here is a list: [pequi, pequi, pequi, pequi, pequi, pequi, pequi, keule, pequi, pequi]. How many times does 'keule' appear on it?}
    
    \texttt{Answer: Let's think step by step. To determine how many times the symbol 'keule' appears in the list [pequi, pequi, pequi, pequi, pequi, pequi, pequi, keule, pequi, pequi], we can simply count the occurrences of 'keule' within the list.}
    \texttt{Looking at the list, we see:}
    \texttt{- There are eight 'pequi' symbols.}
    \texttt{- There is one 'keule' symbol.} 
    \texttt{So, 'keule' appears once in the list.}
    \texttt{So the answer is 1.}
    
    \vspace{0.5em}
    \textcolor{blue}{\texttt{[...7 more examples]}}
    \vspace{0.5em}
    
    \texttt{Instruction: Answer the given question. You will end your response with a sentence in the format of 'So the answer is <number>.'}
    
    \texttt{Question: Here is a list: [keule, pequi, pequi, keule, pequi, keule, keule, pequi, keule, pequi]. How many times does 'keule' appear on it?}
    
    \texttt{Answer:}
    \end{tcolorbox}
    \captionof{table}{Prompt of counting uncommon fruits (CoT setting)}
    \label{tab:prompt_counting_longtail_cot}
\end{table*}

\begin{table*}[htbp]
    \centering
    \begin{tcolorbox}[
        colback=gray!3!white,
        colframe=gray!50!black,
        boxrule=0.5pt,
        width=13cm,
        arc=2mm,
        title=\text{Prompt Template for Capitalization, Base, Direct},
        fonttitle=\bfseries,
        coltitle=white,
    ]
    \texttt{Instruction: Answer the given question. You will end your response with a sentence in the format of `So the answer is \textless string\textgreater.'}
    
    \texttt{Question: Here is a string: "cartoons for victory". Change the format of the string so that it can be a title.}
    
    \texttt{Answer: So the answer is Cartoons for Victory.}
    
    \vspace{0.5em}
    \textcolor{blue}{\texttt{[...Four more examples]}}
    \vspace{0.5em}
    
    \texttt{Instruction: Answer the given question. You will end your response with a sentence in the format of `So the answer is \textless string\textgreater.'}
    
    \texttt{Question: Here is a string: "simple explanation of work ideas". Change the format of the string so that it can be a title.}
    
    \texttt{Answer:}
    \end{tcolorbox}
    \captionof{table}{Prompt of changing title in capitalization task (Direct setting)}
    \label{tab:prompt_cap_title_direct}
\end{table*}

\begin{table*}[htbp]
    \centering
    \begin{tcolorbox}[
        colback=gray!3!white,
        colframe=gray!50!black,
        boxrule=0.5pt,
        width=13cm,
        arc=2mm,
        title=\text{Prompt Template for Capitalization, Longtail, Direct},
        fonttitle=\bfseries,
        coltitle=white,
    ]
    \texttt{Instruction: Answer the given question. You will end your response with a sentence in the format of `So the answer is \textless string\textgreater.'}
    
    \texttt{Question: Here is a string: "cartoons for victory". Change the format of the string so that only the first letter of the last word is capitalized.}
    
    \texttt{Answer: So the answer is cartoons for Victory.}
    
    \vspace{0.5em}
    \textcolor{blue}{\texttt{[...Four more examples]}}
    \vspace{0.5em}
    
    \texttt{Instruction: Answer the given question. You will end your response with a sentence in the format of `So the answer is \textless string\textgreater.'}
    
    \texttt{Question: Here is a string: "simple explanation of work ideas". Change the format of the string so that only the first letter of the last word is capitalized.}
    
    \texttt{Answer:}
    \end{tcolorbox}
    \captionof{table}{Prompt of capitalizing the first letter of the last word in the string (Direct setting)}
    \label{tab:prompt_cap_last_word_direct}
\end{table*}

\begin{table*}[htbp]
    \centering
    \begin{tcolorbox}[
        colback=gray!3!white,
        colframe=gray!50!black,
        boxrule=0.5pt,
        width=13cm,
        arc=2mm,
        title=\text{Prompt Template for Capitalization, Base, CoT},
        fonttitle=\bfseries,
        coltitle=white,
    ]
    \texttt{Instruction: Answer the given question. You will end your response with a sentence in the format of `So the answer is \textless string\textgreater.'}
    
    \texttt{Question: Here is a string: `cartoons for victory'. Change the format of the string so that it can be a title.}
    
    \texttt{Answer: Let's think step by step. To convert the string into a proper title, we need to capitalize the major words, but do not capitalize short conjunctions unless they are the first or last word. We can traverse each word iteratively. `cartoons' becomes `Cartoons'. `for' becomes `for'. `victory' becomes `Victory'. So the answer is Cartoons for Victory.}
    
    \vspace{0.5em}
    \textcolor{blue}{\texttt{[...Four more examples]}}
    \vspace{0.5em}
    
    \texttt{Instruction: Answer the given question. You will end your response with a sentence in the format of `So the answer is \textless string\textgreater.'}
    
    \texttt{Question: Here is a string: the roots of hinduism: the early aryans and the indus civilization'. Change the format of the string so that it can be a title.}
    
    \texttt{Answer:}
    \end{tcolorbox}
    \captionof{table}{Prompt of changing title in capitalization task (CoT setting)}
    \label{tab:prompt_cap_title_cot}
\end{table*}

\begin{table*}[htbp]
    \centering
    \begin{tcolorbox}[
        colback=gray!3!white,
        colframe=gray!50!black,
        boxrule=0.5pt,
        width=13cm,
        arc=2mm,
        title=\text{Prompt Template for Capitalization, Longtail, CoT},
        fonttitle=\bfseries,
        coltitle=white,
    ]
    \texttt{Instruction: Answer the given question. You will end your response with a sentence in the format of `So the answer is \textless string\textgreater.'}
    
    \texttt{Question: Here is a string: "cartoons for victory". Change the format of the string so that only the first letter of the last word is capitalized.}
    
    \texttt{Answer: Let's think step by step. The last word in the string is 'victory', so we capitalize the first letter of it and it becomes 'Victory'. So the answer is cartoons for Victory.}
    
    \vspace{0.5em}
    \textcolor{blue}{\texttt{[...Four more examples]}}
    \vspace{0.5em}
    
    \texttt{Instruction: Answer the given question. You will end your response with a sentence in the format of `So the answer is \textless string\textgreater.'}
    
    \texttt{Question: Here is a string: "the roots of hinduism: the early aryans and the indus civilization". Change the format of the string so that only the first letter of the last word is capitalized.}
    
    \texttt{Answer:}
    \end{tcolorbox}
    \captionof{table}{Prompt of capitalizing the first letter of the last word in the string (CoT setting)}
    \label{tab:prompt_cap_last_cot}
\end{table*}